\documentclass[twoside]{article}
\usepackage{hal,palatino,epsfig,latexsym,natbib}

\usepackage{amsmath,amssymb}
\usepackage{color}
\usepackage{booktabs}
\usepackage{pifont}
\usepackage{mathabx}
\usepackage{url}
\usepackage{array}

\newtheorem{definition}{Definition}

\newcommand{\set}[1]{\left\{{#1}\right\}}

\newcommand{\Nat}{\mathbb{N}}
\newcommand{\Ind}{\mathbb{I}}

\newcommand{\symmetric}{\mathtt{Sym}}
\newcommand{\Neig}[0]{{\mathcal N}}
\newcommand{\A}[0]{\texttt{A}}
\newcommand{\B}[0]{\texttt{B}}
\newcommand{\C}[0]{\texttt{C}}
\newcommand{\D}[0]{\texttt{D}}
\newcommand{\E}[0]{\texttt{E}}
\newcommand{\F}[0]{\texttt{F}}

\newcommand{\hide}[1]{}

\begin{document}

\ecjHeader{x}{x}{xxx-xxx}{201X}{Inventory of Low-Dimensional Invariant Pseudo-Boolean Landscapes}{A. Liefooghe \& S. Verel}

\title{\bf Inventory of the 12\,007\\Low-Dimensional Pseudo-Boolean Landscapes\\Invariant to Rank, Translation, and Rotation}

\author{
       \name{\bf Arnaud Liefooghe} \hfill \addr{arnaud.liefooghe@univ-littoral.fr}\\
        \addr{Univ. Littoral Côte d'Opale, UR 4491, LISIC, F-62100 Calais, France}
\AND
        \name{\bf Sébastien Verel} \hfill \addr{sebastien.verel@univ-littoral.fr}\\
        \addr{Univ. Littoral Côte d'Opale, UR 4491, LISIC, F-62100 Calais, France}
\thanks{~~~The authors contributed equally to this work and are listed in alphabetical order.}
}

\maketitle

\begin{abstract}
Many randomized optimization algorithms are rank-invariant, relying solely on the relative ordering of solutions rather than absolute fitness values. We introduce a stronger notion of \emph{rank landscape invariance}: two problems are equivalent if their ranking, but also their neighborhood structure and symmetries (translation and rotation), induce identical landscapes. This motivates the study of rank landscapes rather than individual functions. While prior work analyzed the rankings of injective function classes in isolation, we provide an exhaustive inventory of the invariant landscape classes for pseudo-Boolean functions of dimensions $1$, $2$, and $3$, including non-injective cases. Our analysis reveals $12\,007$ classes in total, a significant reduction compared to rank-invariance alone. We find that non-injective functions yield far more invariant landscape classes than injective ones. In addition, complex combinations of topological landscape properties and algorithm behaviors emerge, particularly regarding deceptiveness, neutrality, and the performance of hill-climbing strategies. The inventory serves as a resource for pedagogical purposes and benchmark design, offering a foundation for constructing larger problems with controlled hardness and advancing our understanding of landscape difficulty and algorithm performance.
\end{abstract}


\section{Introduction}

Randomized search heuristics are typically applied or evaluated on specific optimization problem instances, yet their performance generalizes across entire classes of problems that are equivalent under certain transformations. Established algorithms such as tabu search, iterated local search, and genetic algorithms with tournament selection are rank-invariant: they rely solely on pairwise comparisons of solutions rather than absolute fitness values~\citep{Whitley89,unanue2024characterization}. For such algorithms, two problems are equivalent if they induce the same ranking of solutions.

While the number of functions is infinite, the number of distinct rankings is finite~\citep{hernando2019characterising}. This enables a systematic enumeration of equivalence classes. This observation has inspired research analyzing the rankings of combinatorial optimization problems as equivalence classes under rank-preserving transformations; see, e.g., ~\citet{ceberio2017we,hernando2019characterising,unanue2024characterization}. This approach explores which problem classes generate identical rankings and includes counting and characterizing the distinct rankings induced by different problem domains~\citep{unanue2023natural,unanue2024characterization}. These studies often leverage connections to Fourier bases, which offer a spectral perspective on problem structure~\citep{Elorza2025}.

An earlier direction in evolutionary computation investigates minimal problem instances that exhibit specific properties, such as deceptiveness. For example, \citet{goldberg1987simple}'s minimal deceptive problem demonstrates how a 2-dimensional pseudo-Boolean function can mislead simple hill climbers while illustrating the role of Walsh coefficients in characterizing deceptiveness. These minimal problems serve as easy-to-understand examples and have been extended to higher dimensions~\citep{mitchell1991royal} or through concatenation of multiple deceptive problems~\citep{ackley1987empirical,deb1993analyzing}, thus providing insights into how optimization challenges scale for algorithms. While minimal problems illustrate specific landscapes, our exhaustive inventory covers all possible low-dimensional landscapes.

However, existing work has focused exclusively on injective functions, where each solution has a distinct fitness value—as many ranks or fitness levels as solutions. This restriction overlooks the rich spectrum of non-injective functions, where ties in fitness values introduce neutrality and plateaus that significantly impact algorithm dynamics~\citep{ReiSta01,Pitzer2012}. More importantly, prior work on rankings considers only the ordering of solutions without accounting for the neighborhood structure. Yet most randomized optimization algorithms depend not only on the ranking but also on how solutions are connected in the landscape. Two problems may have different rankings but share identical structures if their neighborhood relations remain unchanged under landscape transformations.

In this paper, we build upon and extend these earlier efforts by introducing a stronger notion of equivalence: \emph{rank landscape invariance}. In addition to rankings, we consider the neighborhood structure among solutions and their symmetries to determine whether two problems induce identical rank landscapes. These symmetries capture bijective transformations of the search space, such as inverting the meaning of bit values (translations) or reordering variable positions (rotations). They induce identical search dynamics for rank-invariant algorithms. Our approach reveals fundamental structures that prior analyses missed. Additionally, we include non-injective functions, where solutions may share the same rank (including neutral neighbors).

We perform a systematic and exhaustive enumeration of unique rank landscapes and provide the first exhaustive inventory of invariant landscape classes for pseudo-Boolean functions of dimensions $n=1$, $2$, and $3$, including non-injective functions. Our analysis shows that rank landscape invariance significantly reduces the number of unique problems compared to rank-invariance alone---by a factor of $4$ for $n=2$ and $45$ for $n=3$. In addition, we find that non-injective functions yield far more invariant classes than injective ones---$78\%$ for $n=2$ and $92\%$ for $n=3$. We introduce a graph visualization method to highlight and contrast their key topological features. We also go beyond deceptiveness and systematically characterize the structural properties of these landscapes---deceptiveness, but also global and local optimality, neutrality, and plateaus---and their implications for algorithm performance, including the performance of baseline hill-climbing strategies.

Low-dimensional problems ($n \leq 3$) are particularly valuable in this context. While they may appear trivial, they serve as minimal examples illustrating a wide variety of complex landscape analysis concepts. These dimensions represent the smallest non-trivial cases where interactions between variables ($n=2$) and higher-order effects ($n=3$) can be systematically explored. They also provide building blocks for constructing larger problems. For instance, QUBO~\citep{GloverKD19} or SAT~\citep{hoos2000satlib} instances can be decomposed into combinations of smaller subproblems or clauses~\citep{boros2002}, similar to how surrogate meta-models can be defined as weighted sums of kernels~\citep{santin2021kernel}.
Notably, any SAT instance can be reduced to 3-SAT~\citep{Karp1972}, with clauses of size $n=3$.
Understanding subproblem properties is therefore essential for benchmark and algorithm design. Our inventory provides a complete catalog of small subproblem topologies---analogous to basis functions in Fourier analysis---and offers a resource for constructing larger problems with controlled difficulty. We hope it will serve as an open resource for benchmarking and improving explainability in landscape analysis.

The paper is organized as follows. Section~\ref{sec:rank_inv} introduces rank invariance. Section~\ref{sec:landscape_inv} formalizes landscape invariance and describes our methodology for enumerating invariant classes. Section~\ref{sec:summary} illustrates the key concepts through examples. Section~\ref{sec:inventory} presents the complete inventory, including counts of invariant landscape classes, characterization of landscape properties, and algorithm insights. Finally, Section~\ref{sec:conclu} discusses implications for future research.

\section{Method: Rank Invariance}
\label{sec:rank_inv}
In this section, we define the rank-invariance property and establishes equivalence classes within the set of (pseudo-Boolean) fitness functions.

\subsection{Rank-Invariant Functions}
Let us consider a fitness function $f : X \to Y$ defined over the domain $X$  and mapping to a codomain $f(X) = Y$. 
Both $X$ and $Y$ are assumed to be countable sets.
Without loss of generality, we assume the fitness function $f$ is to be minimized.
The concepts in this section apply to all function classes.
For pseudo-Boolean optimization specifically, the domain $X = \set{0,1}^n$ is the set of binary strings of length $n$, with a search space size of $\vert X \vert = 2^n$.

We are interested in the ranking of solutions in set $X$ induced by $f$.
The \emph{image} of~$f$ is the set of all elements in $Y$ (without repetition) that are mapped to elements in~$X$:%
\begin{equation}
I(f) = \set{ f(x) \mid x \in X }
\end{equation}
The cardinality of this image $k = \vert I(f) \vert$ represents the number of distinct fitness values that $f$ produces.
When $f$ is injective, $k = \vert X \vert$. 
However, in contrast with~\citet{unanue2023natural}, we are also interested in functions that may have fewer than $\vert X \vert$ fitness values, that is, problems where different solutions yield equal fitness values.
In the more general case, $1 \leqslant k \leqslant \vert X \vert$. 

Let $\set{y_1, y_2, \ldots, y_k}$ be the set of distinct elements in $I(f)$ ordered such that $y_1 < y_2 < \ldots < y_k$.
This is the sorted list of unique values in the image of $f$.
We now define the \emph{rank function} $r_f : X \to \Nat$ based on the ordered elements in the image of $f$:
\begin{equation}
r_f(x) = \sum_{i=1}^{k} i \cdot \Ind(f(x) = y_i)
\end{equation}
where $\Ind(f(x) = y_i)$ is an indicator function that equals $1$ if $f(x) = y_i$ and $0$ otherwise.
For any $x \in X$, $r_f(x)$ gives the position of $f(x)$ in the sorted list of unique values in $I(f)$.
This maps each solution to a rank (or fitness level), based on the position of its fitness value in the sorted sequence. 
To avoid confusion with fitness values, we map solution ranks from $1$ to $k$ using letters in the alphabet: \A\ for rank $1$, \B\ for rank $2$, and so on.
Following this principle, global optima have a rank of $A$.
Table~\ref{tab:rank_exp} shows how solutions are ranked for example pseudo-Boolean fitness functions with $n=2$ binary variables.

\begin{table}[t!]
\caption{Solutions, fitness values, and corresponding ranks for five example two-dimensional pseudo-Boolean functions to be minimized.\smallskip}
\label{tab:rank_exp}
\centering%
\begin{tabular}{c|cc|cc|cc|cc|cc}
\toprule
$x$		&	$f_1(x)$	&	rank	&	$f_2(x)$	&	rank	&	$f_3(x)$	&	rank	&	$f_4(x)$	&	rank	&	$f_5(x)$	&	rank	\\
\midrule
\texttt{00}	&	$4.0$	&	\C	&	$3.6$	&	\C	&	$7.0$	&	\D	&	$7.0$	&	\D	&	$3.0$	&	\B	\\
\texttt{01}	&	$1.0$	&	\A	&	$2.9$	&	\A	&	$2.9$	&	\B	&	$3.0$	&	\C	&	$3.0$	&	\B	\\
\texttt{10}	&	$9.0$	&	\D	&	$3.7$	&	\D	&	$3.0$	&	\C	&	$2.9$	&	\B	&	$7.0$	&	\C	\\
\texttt{11}	&	$3.0$	&	\B	&	$3.1$	&	\B	&	$2.0$	&	\A	&	$2.0$	&	\A	&	$2.0$	&	\A	\\
\bottomrule
\end{tabular}
\end{table}

\begin{definition}[Rank-invariant functions]
Two functions $f_1$, $f_2$ are rank-invariant if and only if, $\forall x \in X$, $r_{f_1}(x) = r_{f_2}(x)$.
\end{definition}
From the examples in Table~\ref{tab:rank_exp}, only $f_1$ and $f_2$ are rank-invariant.

Rank-invariant functions share similar properties and exhibit the same global structure.
Therefore, it is generally expected that search algorithms behave consistently when applied to them.
Indeed, many optimization algorithms rely solely on how solutions rank relative to each other. 

\subsection{Rank-Invariant Algorithms}

Local search and evolutionary computation algorithms can be modeled as a function that generates a sequence of solutions (or trajectory) based on a fitness function~\mbox{$f : X \to Y$}.
Though these algorithms are often inherently stochastic, the produced solution sequence becomes deterministic when given a fixed sequence of pseudo-random numbers $R$.
With identical initial conditions and random sequences, a rank-invariant algorithm $A$ will consistently produce the same solution sequences  $A(f, R) = (x^{(1)}, \ldots, x^{(\ell)})$, thus ensuring reproducible results%
\footnote{For population-based algorithms, a sequence can be defined in terms of multisets of solutions rather than individual solutions.}.
%
\begin{definition}[Rank-invariant algorithm]
An algorithm $A$ is rank-invariant if and only if, for any two rank-invariant fitness functions $f_1$ and $f_2$ and a given sequence of pseudo-random numbers $R$, the sequence of solutions is the same: $A(f_1, R) = A(f_2, R)$.
\end{definition}
Rank-invariant algorithms maintain consistent behavior across different problems with the same solution ordering.
These algorithms rely solely on the relative ranking of solutions, not their actual fitness values.
While the sequence of solutions produced by the algorithm depends on both the random sequence and the problem structure, rank-invariant algorithms perform consistently across rank-invariant functions.
This makes them invariant to any strictly monotonic transformation of the fitness function.

A $(1+1)$-EA~\citep{Rudolph98} is rank-invariant as it simply compares and selects the best between the parent and the offspring.
Similarly, a hill-climbing algorithm~\citep{HooStu05sls-mk} is rank-invariant if it selects the best neighbor and breaks ties at random.
Standard tabu search~\citep{GloLag97} also maintains rank invariance, while simulated annealing~\citep{NikJac2010} does \emph{not}: it relies on actual fitness values to calculate acceptance probabilities for candidate solutions.

\subsection{Count of Rank-Invariant Pseudo-Boolean Function Classes}
\label{sec:count_rank_invariant_functions}

We now examine the combinatorics of possible rankings induced by the family of (small-size) pseudo-Boolean functions.
Indeed, while the number of possible fitness functions is infinite, rank-invariant functions form a finite set~\citep{hernando2019characterising}.
We enumerate classes of rank-invariant functions with $1 \leqslant k \leqslant \vert X \vert$ distinct fitness values.
Specifically, we examine functions~$f$ such that $k = |f(X)|$ is the size of the image of $f$, denoted as $I(f)$.%

\paragraph{Partitions.}%
We recall that $\set{y_1, \ldots, y_i, \ldots, y_k}$ is the set of distinct elements in $I(f)$ ordered such that $y_1 < \ldots < y_i < \ldots < y_k$.
For all $i \in \set{1, \ldots,k}$, let $\lambda_i$ denote the number of solutions with rank $i$---that is, the number of solutions $x \in X$ such that $f(x)=y_i$.
Note that $\sum_{i=1}^k \lambda_i = \vert X \vert$.
The vector $\lambda = (\lambda_1, \ldots, \lambda_k)$ represents a partition of $\vert X \vert$ into $k$ parts, where each $\lambda_i \geqslant 1$.
For injective functions, with $k = \vert X \vert = 2^n$ different values, there is only one partition $\lambda = (1, \ldots, 1)$ of size $2^n$.

We first enumerate the number of possible partitions $(\lambda_1, \ldots, \lambda_k)$ with $k$ distinct ranks.
We use $p_k$ to denote the number of partitions containing $k$ ranks.
To ensure that each of the $k$ ranks contains at least one element (i.e., $\lambda_i \geqslant 1$), we decompose the $k$-dimensional partition into
$\lambda = (1, \ldots, 1) + (\lambda_1 - 1, \ldots, \lambda_k - 1)$.
Table~\ref{tab:partitions_n2} shows all partitions for pseudo-Boolean functions with $n=2$ variables (column $p_k$).

\begin{table}[t!]
\caption{Enumeration of partitions for two-dimensional pseudo-Boolean functions: $p_k$ denotes the number of partitions with $k$ ranks, $F_\lambda$ denotes the number of rank-invariant function classes for partition $\lambda$.\smallskip}
\label{tab:partitions_n2}
\centering%
\begin{tabular}{l|lllll|c|r}
\toprule
		&	$(1)^k$		&	$+$	&	$i$			&	$=$	&	$\lambda$		&	$p_k$	&	$F_{\lambda}$	\\
\midrule
$k=4$	&	$(1,1,1,1)$	&	$+$	&	$(0,0,0,0)$	&	$=$	&	$(1,1,1,1)$	&	$1$		&	$24$			\\
\midrule
$k=3$	&	$(1,1,1)$		&	$+$	&	$(1,0,0)$		&	$=$	&	$(2,1,1)$		&	$3$		&	$12$			\\
		&				&	$+$	&	$(0,1,0)$		&	$=$	&	$(1,2,1)$		&			&	$12$			\\
		&				&	$+$	&	$(0,0,1)$		&	$=$	&	$(1,1,2)$		&			&	$12$			\\
\midrule
$k=2$	&	$(1,1)$		&	$+$	&	$(2,0)$		&	$=$	&	 $(3,1)$		&	$3$		&	$4$			\\
		&				&	$+$	&	$(1,1)$		&	$=$	&	 $(2,2)$		&			&	$6$			\\
		&				&	$+$	&	$(0,2)$		&	$=$	&	 $(1,3)$		&			&	$4$			\\
\midrule
$k=1$	& $(1)$			&	$+$	&	$(3)$			&	$=$	&	 $(4)$		&	$1$		&	$1$			\\
\midrule
\multicolumn{6}{l|}{\textbf{total}}												&	$\mathbf{8}$	&	$\mathbf{75}$	\\
\bottomrule
\end{tabular}
\end{table}

To enumerate all partitions $\lambda$, we must list all possible vectors $\lambda^{\prime} = \lambda - (1)^k$ of dimension~$k$ containing $u = \vert X \vert - k$ unconstrained elements (i.e., each component $\lambda^{\prime}_j$ can equal $0$).
Consequently, there are $p_k = {\vert X \vert - 1 \choose k - 1}$ partitions with $k$ ranks.
This enumeration is equivalent to the dual enumeration of vectors $v = (v_1, \ldots, v_u)$ of dimension~$u$, where each component $v_j \in \set{1, \ldots ,k}$ indicates the position of element $j$ in the partition vector $\lambda^{\prime}$.
Partitions with $k$ ranks can therefore be enumerated through the following iterative process:
\begin{equation}\nonumber%
\text{next}(v) =
\begin{cases}
(v_1, v_2, \ldots\ldots\ldots\ldots\ldots\ldots\ldots, v_u + 1) & \text{if } v_u < k, \\
(v_1, v_2, \ldots, v_j + 1, v_j + 1, \, \ldots, v_j + 1) & \text{otherwise, with } j = \max \{ j \mid v_j < k \}
\end{cases}
\end{equation}

\paragraph{Functions.}%
For a given partition $\lambda$, we enumerate all rank-invariant functions by assigning $\lambda_1$ values to $y_1$, $\lambda_2$ values to $y_2$, and so forth.
This is equivalent to generating anagrams using a bag of $k$ distinct letters, where each letter $y_i$ is duplicated $\lambda_i$ times.
For a given partition~$\lambda$, the number of different functions $F_\lambda$ is:
\begin{equation}
F_\lambda = \frac{2^n!}{\prod_{j=1}^{k} \lambda_j !}
\end{equation}
As pointed out to us by \citet{personal_francis}, the total number of possible rankings corresponds to Fubini numbers---also known as ordered Bell numbers or preferential arrangements---which count the number of ways $m$ competitors can be ranked in a competition, allowing for ties~\citep{oeisA000670}.

Table~\ref{tab:partitions} reports the number of partitions $p_k$ with $k$ ranks, along with their corresponding function counts for $n=3$ and $n=4$.
The total function count grows drastically with dimension: $75$ for $n=2$ (Table~\ref{tab:partitions_n2}), $545\,835$ for $n=3$, and over $1\,000$ trillion for $n=4$.
This growth rate surpasses that of a factorial function as dimension increases.
In the following, we thus focus on functions of dimensions $1$, $2$, and $3$.

\begin{table}[t!]
\caption{Number of partitions and rank-invariant function classes by rank count $k$ for three- and four-dimensional pseudo-Boolean functions.\smallskip}
\label{tab:partitions}
\centering%
\begin{tabular}[t]{l|r|r}
\multicolumn{3}{c}{${n=3}$}\\
\toprule
		&	$p_\lambda$	&	$\sum_{\lambda : d_\lambda = k} F_{\lambda}$	\\
\midrule
$k=1$	&	$1$	&	$1$			\\
$k=2$	&	$7$	&	$254$		\\
$k=3$	&	$21$	&	$5\,796$		\\
$k=4$	&	$35$	&	$40\,824$		\\
$k=5$	&	$35$	&	$126\,000$	\\
$k=6$	&	$21$	&	$191\,520$	\\
$k=7$	&	$7$	&	$141\,120$	\\
$k=8$	&	$1$	&	$40\,320$		\\
\midrule
\textbf{total}	&	$\mathbf{2^{7} = 128}$	&	$\mathbf{545\,835}$ \\ 
\bottomrule
\end{tabular}
\hfill%
\begin{tabular}[t]{l|r|r}
\multicolumn{3}{c}{${n=4}$}\\
\toprule
		&	$p_\lambda$	&	$\sum_{\lambda : d_\lambda = k} F_{\lambda}$	\\
\midrule
$k=1$	&	$1$		&	$1$				\\
$k=2$	&	$15$		&	$65\,534$				\\
$k=3$	&	$105$	&	$42\,850\,116$			\\
$k=4$	&	$455$	&	$4\,123\,173\,624$			\\
$k=5$	&	$1\,365$	&	$131\,542\,866\,000$		\\
$k=6$	&	$3\,003$	&	$1\,969\,147\,121\,760$		\\
$k=7$	&	$5\,005$	&	$16\,540\,688\,324\,160$		\\
$k=8$	&	$6\,435$	&	$86\,355\,926\,616\,960$		\\
$k=9$	&	$6\,435$	&	$297\,846\,188\,640\,000$	\\
$k=10$	&	$5\,005$	&	$703\,098\,107\,712\,000$	\\
$k=11$	&	$3\,003$	&	$1\,155\,068\,769\,254\,400$	\\
$k=12$	&	$1\,365$	&	$1\,320\,663\,933\,388\,800$	\\
$k=13$	&	$455$	&	$1\,031\,319\,184\,896\,000$	\\
$k=14$	&	$105$	&	$524\,813\,313\,024\,000$	\\
$k=15$	&	$15$		&	$156\,920\,924\,160\,000$	\\
$k=16$	&	$1$		&	$20\,922\,789\,888\,000$		\\
\hline
\textbf{total}	&	$\mathbf{2^{15} = 32\,768}$	&	$\approx \mathbf{2^{52.23}}$	\\ 
\hline
\end{tabular}
\end{table}

\section{Method: Landscape Invariance}
\label{sec:landscape_inv}
We are not only interested in rank-invariant fitness functions but also in their underlying landscapes and in the core landscape-invariant problems in pseudo-Boolean optimization.
This section defines landscape invariance and examine low-dimensional pseudo-Boolean invariant landscapes.

\subsection{Landscape-Invariant Transformations: Translations and Rotations}

A fitness landscape consists of a triplet $\mathcal{L}=(X, \Neig, f)$ such that $X = \set{0,1}^n$ is the search space, $\Neig$ is the neighborhood relation between solutions, and $f$ is the fitness function~\citep{merz2004}.
Following established practice in pseudo-Boolean optimization~\citep{HooStu05sls-mk}, we define the neighborhood relation such that two solutions are neighbors when they differ by exactly one bit --- that is, when they have a Hamming distance of $1$.
This neighborhood is symmetric: for any $x, x^\prime \in X$, $x^\prime \in \Neig(x) \iff x \in \Neig(x^\prime)$.
Each solution has $n$ neighbors. 

We define the landscape as
a \emph{graph} $G_\mathcal{L}(X,E)$ such that nodes $X$ are solutions, and edges $E$ represent neighborhood connections~\citep{Stadler1995landscapes}. 
That is, there is an edge between $x$ and $x^\prime \in X$ if and only if $x^\prime \in \Neig(x)$.
Each node $x \in X$ is assigned a value that represents the fitness of the corresponding solution $f(x)$.
This graph depicts the landscape that algorithms explore under the $1$-flip operator.
It is an $n$-dimensional hypercube, with $2^n$ nodes and \mbox{$n \cdot 2^{n-1}$} edges.

The symmetric group $\symmetric(X)$ represents all possible permutations of these nodes, with order $2^n !$ --- through not all permutations qualify as graph automorphisms.
The landscape graph $G_\mathcal{L}$ has a specific group of automorphisms, totaling $2^n \cdot n !$ in number~\citep{gregor2012}.
The automorphism group is a subgroup of the symmetric group on the nodes set that preserves the graph adjacency.
These automorphisms can be described as follows:
\begin{description}
\item[Translations:]
A translation adds a fixed binary string $z \in X$ to every node (solution), mapping each node $x \in X$ to $t_z(x) = x \oplus z$ where $\oplus$ represents the bitwise \texttt{XOR} operation.
For any node $x \in X$ and its neighbor $x^\prime \in \Neig(x)$, the translation yields $t_z(x) = x \oplus z$ and $t_z(x^\prime) = x^\prime \oplus z$ respectively, maintaining graph adjacency (neighborhood connections).
The total number of possible translations equals $2^n$, corresponding to all possible binary strings of length $n$.
\item[Rotations:]
A rotation rearranges the bits in each node (solution).
Given a permutation $\sigma \in \symmetric(X)$, the rotation $r_\sigma$ transforms each node $x = (x_1, x_2, \ldots, x_n) \in X$ to $r_\sigma(x)= (x_{\sigma(1)}, x_{\sigma(2)}, \ldots, x_{\sigma(n)})$ while preserving adjacency.
There are $n !$ possible rotations, corresponding to the number of ways to permute $n$ elements.
\end{description}
In simple terms, a translation corresponds to an inversion of the interpretation of some variable values ($\texttt{0} \leftrightarrow \texttt{1}$), while a rotation corresponds to a reordering of the positions of variables.
The complete set of automorphisms of the hypercube is composed with a translation and a rotation~\citep{gregor2012}.
This yields a total of $2n \cdot n !$ automorphisms.
To illustrate these concepts, we examine low-dimensional hypercubes:
\begin{itemize}
\item For $n=1$, there are $2^1 \cdot 1 ! = 2$ automorphisms, consisting of $2$~translations and $1$~rotation.
\item For $n=2$, there are $2^2 \cdot 2 ! = 8$ automorphisms, consisting of $4$~translations and $2$~rotations (Table~\ref{tab:trans_n2} in Appendix).
\item For $n=3$, there are $2^3 \cdot 3 ! =  48$ automorphisms, consisting of $8$~translations and $6$~rotations (Table~\ref{tab:trans_n3} in Appendix).
\end{itemize}
The automorphism group of the hypercube of dimension $n$ exhibits a rich structure, reflecting the high degree of symmetry in these landscape graphs.

\subsection{Invariant Landscapes}
We say that two landscapes are graph-invariant if there exists a bijective mapping between their elements that preserves the neighborhood structure defined by $\Neig$.%
\begin{definition}[Invariant landscapes]
Two landscapes \mbox{$\mathcal{L}_1=(X, \Neig, f_1)$} and \mbox{$\mathcal{L}_2=(X, \Neig, f_2)$} are graph-invariant if and only if there exists a transformation $\tau : X \to X$ such that:
\begin{enumerate}
\item $\tau$ is bijective, meaning it belongs to the symmetric group $\symmetric(X)$;%
\item $\tau$ preserves the neighborhood structure defined by $\Neig$: $\forall x \in X$, $\tau( \Neig(x) ) = \Neig( \tau(x) )$;%
\item $f_1$ and $f_2 \circ \tau$ are rank-invariant: $\forall x \in X$, $r_{f_1}(x) = r_{f_{2} \circ \tau}(x)$ where $\circ$ defines the composition $(f_{2} \circ \tau)(x) = f_{2}(\tau(x))$.
\end{enumerate}
\end{definition}
In this case, we say that $\mathcal{L}_2$ is a transformation of $\mathcal{L}_1$ by $\tau$: $\mathcal{L}_2 = \tau(\mathcal{L}_1)$.
Invariant landscapes preserve their global structure: up to transformation $\tau$, each solution node maintains the same rank and the same set of neighbors.

Returning to the examples from Table~\ref{tab:rank_exp}, we previously noted that functions $f_1$ and $f_2$ are rank-invariant, which naturally makes them landscape-invariant. When examining the other functions, we find that $f_3$ transforms into $f_1$ through a translation $t(x) = x \oplus z$ with $z=10$ (inverting the left-hand bit), while $f_4$ can be converted into 
$f_3$ by reordering (swapping) both variables.
These simple transformations show that $f_1$, $f_2$, $f_3$, and $f_4$ all belong to the same landscape invariance class. By contrast, $f_5$ belongs to a different class, as evidenced by its fewer unique ranks.

\subsection{Landscape-Invariant Algorithms}
Building on the notion of graph-invariant landscapes, we define landscape-invariance for search algorithms as follows:

\begin{definition}[Landscape-invariant algorithm]
An algorithm $A$ is landscape-invariant if and only if, for any two graph-invariant landscapes $\mathcal{L}_1$ and $\mathcal{L}_2$ related by a transformation $\tau$ (i.e., $\mathcal{L}_1 = \tau(\mathcal{L}_2)$), and for any fixed sequence of pseudo-random numbers $R$, the sequences of solutions visited by $A$ on $\mathcal{L}_1$ and $\mathcal{L}_2$ are identical up to the transformation $\tau$. That is, if $S_1$ and $S_2$ are the solution sequences visited by $A$ on $\mathcal{L}_1$ and $\mathcal{L}_2$ respectively, then $S_1 = \tau(S_2)$.
\end{definition}

This definition accounts for the algorithm's dependence on the initial solution, even in the simple case of deterministic algorithms. The initial solution is typically sampled uniformly across the search space, and our interest is in the distribution of performance when starting from any point in that space.

For example, a hill-climbing algorithm is \emph{not} landscape-invariant when it breaks ties between neighbors based on a predefined order, as the sequence of visited solutions would then depend on a specific labeling of the landscape. However, it \emph{is} landscape-invariant if ties are broken at random, since the solution sequence remains consistent up to the transformation $\tau$. Evolutionary algorithms with tournament selection~\citep{Whitley89} and representation-agnostic operators are also landscape-invariant, as they depend only on fitness comparisons and stochasticity. In contrast, algorithms exploiting problem-specific rules, such as tailored crossover, may violate landscape-invariance.
Landscape-invariance ensures that an algorithm's performance is consistent across invariant landscapes, enabling robust theoretical analysis and fair benchmarking.

\subsection{Count of Invariant Landscape Classes}
In this section, we calculate the number of invariant landscapes for pseudo-Boolean fitness functions.
We first examine injective landscapes (with no equivalent solutions), then analyze landscapes with ranks ranging from 1 to $\vert X \vert$.

\paragraph{Landscapes of Injective Functions.}
We begin by examining landscapes with injective fitness functions, where each fitness value is unique and the order of the codomain is $k = \vert f(X) \vert = \vert X \vert = 2^n$.
With $2^n$ distinct elements, where each permutation creates a unique ordering, we obtain $2^n!$ possible rankings.
The number of invariant landscapes corresponds to the number of possible rankings divided by the number of automorphisms:%
\begin{equation}
\label{eq:n_auto_injective}
\frac{2^n!}{(2^n \cdot n!)} = \frac{(2^n - 1)!}{n!}
\end{equation}
Table \ref{tab:injective} reports the number of distinct classes of invariant landscapes with $n \leqslant 4$ binary variables.
While the count is reasonable up to $n=3$, the number of classes for $n=4$ surpasses ten billion, making enumeration impractical.

\begin{table}[t!]
\caption{Number of invariant landscape classes for \emph{injective} pseudo-Boolean fitness functions with $n \leqslant 4$ variables.\smallskip}
\label{tab:injective}
\centering%
\begin{tabular}{l|r|r|r|r}
\toprule
		&	\textbf{ranks}	&	\textbf{rankings}					&	\textbf{automorphisms}	&	\textbf{invariant landscape classes}		\\
		&	$2^n$		&	$2^n!$							&	$2^n \cdot n!$			&	$(2^n - 1)! / n!$						\\
\midrule
$n=1$	&	$2$			&	$2$								&	$2$					&	$1$								\\
$n=2$	&	$4$			&	$24$								&	$8$					&	$3$								\\
$n=3$	&	$8$			&	$40\,320$							&	$48$					&	$840$							\\
$n=4$	&	$16$			&	$\approx 2^{44.25} \approx 10^{13.32}$	&	$384$				&	$\approx 2^{35.67} \approx 10^{10.73}$	\\
\bottomrule
\end{tabular}
\end{table}

\paragraph{Landscapes of Functions with $k$ Ranks.}%
For injective fitness functions, each transformation $\tau$ forms a bijection within the symmetric group $\symmetric(X)$.
Any two transformations $\tau$ and $\tau^\prime$ produce distinct landscapes.
For any graph-invariant transformation $\tau$, the landscapes $\mathcal{L}$ and $\tau(\mathcal{L})$ are different.
However, when the order of the codomain $k$ falls below $\vert X \vert$, there are solutions with equal fitness.
The internal symmetries induced by the ranking thus reduce the number of distinct landscapes under graph-invariant transformations.
The simplest example is the constant fitness function, with a single rank ($k=1$).
Here, the landscape remains unchanged regardless of the graph-invariant transformation, yielding a class of a single landscape.
This is to contrast with $\frac{(2^n - 1)!}{n!}$ for injective functions (Eq.~(\ref{eq:n_auto_injective})).%

In order to determine equivalence classes, we computationally enumerate all possible landscapes with $1 \leqslant k \leqslant \vert X \vert$ ranks and calculate their automorphisms using graph-invariant transformations.
This approach efficiently identifies classes for dimensions $1$, $2$, and $3$ within seconds.
However, for $n = 4$, calculations are not feasible due to the vast number of landscapes.
We present and discuss the resulting invariant landscape classes in detail in Section~\ref{sec:inventory}.

\section{Summary of Concepts through Illustrative Examples}
\label{sec:summary}

\begin{figure*}[p]
\vspace{-0.5in}%
\centering%
\includegraphics[width=\columnwidth]{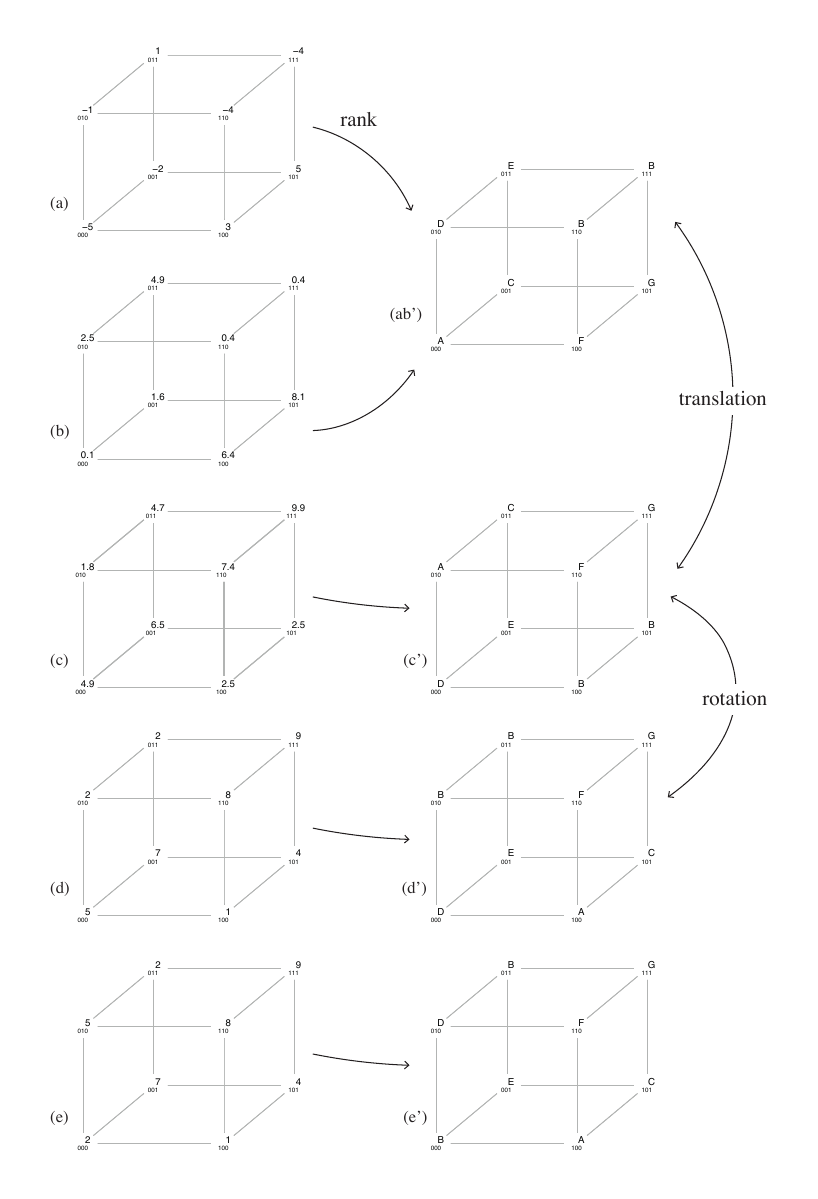}%
\vspace{-0.1in}%
\caption{Fitness and rank landscapes of five three-dimensional pseudo-Boolean function examples, mapping to two distinct invariant landscape classes: (ab$^\prime$), (c$^\prime$), and (d$^\prime$) are automorphic (id \texttt{1580} in the 3D inventory in Section~\ref{sec:inventory}), while (e$^\prime$) belongs to a separate class (id \texttt{1371}).}
\label{fig:summary_example}
\end{figure*}

Before presenting the inventory of invariant landscape classes, we summarize and illustrate the key concepts introduced above through three-dimensional pseudo-Boolean function examples. 
The left side of Figure~\ref{fig:summary_example} displays fitness landscapes for five hypothetical functions to be minimized. In the hypercube graph, each node represents one of the $2^n = 8$ solutions, labeled with its bitstring and fitness value. Edges connect nodes whose solutions are 1-flip neighbors.

For each fitness landscape, we rank the fitness values and construct a rank landscape (shown on the right side of the figure). At this stage, we observe that, despite having quite different fitness distributions, functions (a) and (b) are rank-invariant and map to the same rank landscape~(ab$^\prime$).
Interestingly, functions (c) and (d) exhibit automorphic rank landscapes compared to the one above. The rank landscapes (c$^\prime$) can be obtained from (ab$^\prime$) by flipping the second bit---a translation $(x_1, x_2, x_3) \mapsto (x_1, \neg x_2, x_3)$ such that, for each $x \in X$, $t(x) = x \oplus \texttt{010}$. Similarly, the rank landscape~(d$^\prime$) can be derived from (c$^\prime$) through a $90^{\circ}$ rotation around the $x_3$-axis (reordering bits one and two while fixing the third bit): $(x_1, x_2, x_3) \mapsto (x_2, x_1, x_3)$.
This rotation cycles nodes as follows:
$\texttt{010} \mapsto \texttt{100}$, $\texttt{011} \mapsto \texttt{101}$, $\texttt{100} \mapsto \texttt{010}$, $\texttt{101} \mapsto \texttt{011}$, while keeping the remaining nodes unchanged.
Therefore, we can transform rank landscape~(ab$^\prime$) into rank landscape~(d$^\prime$) by combining a $90^{\circ}$ rotation with a one-bit translation: $(x_1, x_2, x_3) \mapsto (\neg x_2, x_1, x_3)$.
These transformations maintain the graph's original structure and neighborhood relationships. Such operations---translations, rotations, and their combinations---are hypercube automorphisms that preserve the rank landscape structure.
Consequently, the landscapes induced by functions (a) through (d) all belong to the same landscape invariance class.

Examining example (e), we notice it appears similar to example (d) at first glance, with only two fitness values (and thus ranks) swapped between solutions \texttt{000} and \texttt{010}. Despite this seemingly minor difference, no automorphism can transform (d$^\prime$) into (e$^\prime$), meaning these two functions actually belong to different invariant landscape classes.
When considering how these landscapes affect search algorithms, it is easy to understand why. While both are deceptive and share the same number of ranks, landscape~(d$^\prime$) contains two adjacent local optima, forming a closed plateau with rank~\B. In contrast, landscape~(e$^\prime$) only has a single local optimum---its second \B-rank node does not qualify as a local optimum because it neighbors a node of rank \A.
This structural difference leads to distinct behaviors even in basic local search algorithms. In landscape~(d$^\prime$), a best-improvement hill-climber has a $50\%$ chance~($4/8$) of finding the global optimum, depending on the initial solution.
Starting from solutions with ranks \A, \C, \D, or \F\ leads to rank \A, while the others lead to a suboptimum (rank \B).
In landscape~(e$^\prime$), however, this probability increases to $62.5\%$~($5/8$), as there is an equal chance of reaching either rank \A\ or \B\ when starting from solutions with ranks \B, \D, or \E.
This illustrates why distinguishing between these two invariant landscape classes is critical for carefully understanding algorithm behavior.

\hide{
\begin{tabular}{c|l|l}
from        &   (d')    &   (e')    \\
\hline
\hline
$A$ &   $A$                     &   $A$                     \\
\hline
$B$ &   $B$                     &   $B$                     \\
    &   $B$                     &   $B\mapsto A$            \\
\hline
$C$ &   $C \mapsto A$           &   $C \mapsto A$           \\
\hline
$D$ &   $D \mapsto A$           &   $D \mapsto B$           \\
    &                           &   $D \mapsto B \mapsto A$ \\
\hline
$E$ &   $E \mapsto B$           &   $E \mapsto B$           \\
    &                           &   $E \mapsto B \mapsto A$ \\
\hline
$F$ &   $F \mapsto A$           &   $F \mapsto A$           \\
\hline
$G$ &   $G \mapsto B$           &   $G \mapsto B$           \\
\hline
\end{tabular}

\bigskip

\begin{tabular}{c|l|l}
from        &   (d')    &   (e')    \\
\hline
\hline
$A$ ($\times 1$)    &   $A$ ($1.0$) &   $A$ ($1.0$)                 \\
$B$ ($\times 2$)    &   $B$ ($1.0$) &   $A$ ($0.5$) or $B$ ($0.5$)  \\
$C$ ($\times 1$)    &   $A$ ($1.0$) &   $A$ ($1.0$)                 \\
$D$ ($\times 1$)    &   $A$ ($1.0$) &   $A$ ($0.5$) or $B$ ($0.5$)  \\
$E$ ($\times 1$)    &   $B$ ($1.0$) &   $A$ ($0.5$) or $B$ ($0.5$)  \\
$F$ ($\times 1$)    &   $A$ ($1.0$) &   $A$ ($1.0$)                 \\
$G$ ($\times 1$)    &   $B$ ($1.0$) &   $B$ ($1.0$)                 \\
\hline
\end{tabular}
}

\section{Inventory of Invariant Landscape Classes}
\label{sec:inventory}

With the fundamental concept of invariant landscapes now explained, we turn our attention to the comprehensive inventory of invariant landscape classes for injective and non-injective pseudo-Boolean functions in this section.
We begin with the complete inventory of one- and two-dimensional landscape classes, followed by selected three-dimensional landscape classes with summary statistics.

\subsection{1D Landscapes}

In 1D landscapes, we have two solutions ($\texttt{0}$ and $\texttt{1}$) with three possible rankings: either the first solution is better $(r_f(0), r_f(1))=$ (\A, \B), the second solution is better $(r_f(0), r_f(1))=$ (\B, \A), or both solutions have the same rank $(r_f(0), r_f(1))=$ (\A, \A). However, since the two solutions are 1-flip neighbors, the distinction between (\A, \B) and (\B, \A) becomes irrelevant: their rank landscapes are automorphic by translation and belong to the same invariant landscape class.

Therefore, for $n=1$, we identified three possible rank landscapes and \textbf{two one-dimensional  invariant landscape classes}, illustrated in Figure~\ref{fig:landscapes1}.
In our visualization, ranks appear in shaded blue, with darker shades indicating better ranks. Global optima are highlighted in yellow---in the 1D case, one class features a plateau of global optima while the other has a single optimum. Edges are directed to show improving moves, with lighter edges conveying neutral moves.
For compactness in the plots, we also simplify the notation (\A, \B), (\B, \A) and (\A, \A) by omitting the parenthesis, giving \A\B,~\B\A~and \A\A, respectively.

\newcommand{\mysizeh}{1in}
\begin{figure*}[t!]
\vspace{-0.2in}
\centering%
\includegraphics[width=.29\columnwidth]{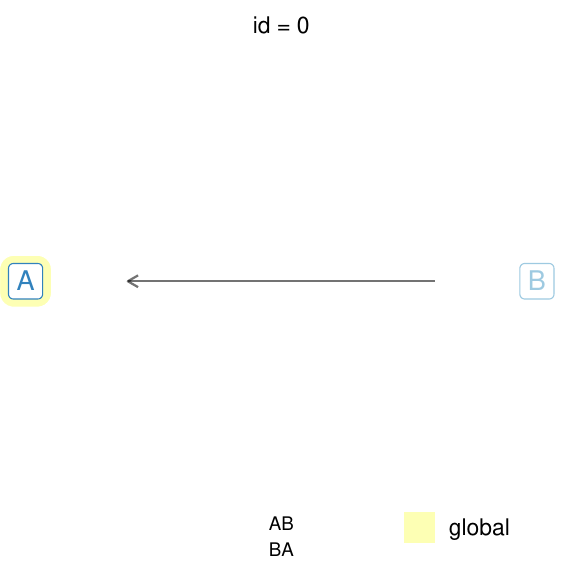}%
\hspace{\mysizeh}%
\includegraphics[width=.29\columnwidth]{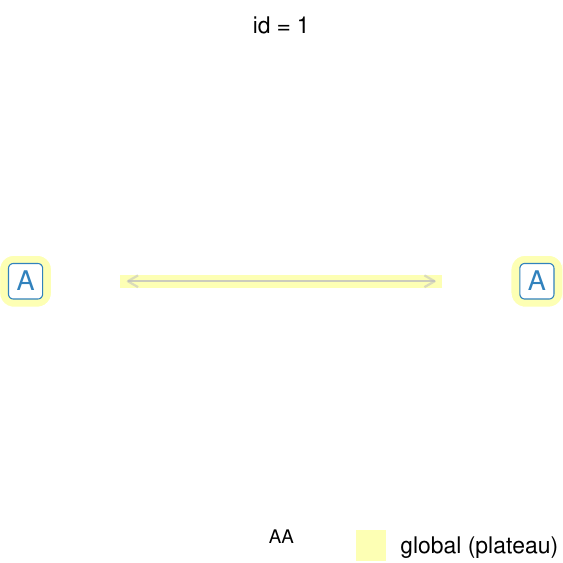}%
\caption{The 2 one-dimensional invariant landscape classes.}
\label{fig:landscapes1}
\end{figure*}

From a ranking perspective, these landscapes both represent Boolean functions where each solution has either rank \A\ (satisfied) or rank \B\ (unsatisfied). Notably, the landscape with distinct ranks (ID \texttt{0}) is falsifiable and corresponds to a 1-SAT clause.
The landscape class with a single rank (ID \texttt{1}) corresponds to the class of constant functions.
Even in this elementary case, we remark that there are as many injective landscape classes as non-injective ones.

From a search optimization perspective, these two landscapes exhibit fundamentally different behaviors. Obviously, there is no deceptive one-dimensional landscape. However, in one class, all solutions share the same rank, creating a degenerate landscape where every solution is a global optimum---making the problem solvable by random sampling. The other class features a single global optimum with minimal search challenge: the optimum can be reached from any starting point in just one step (one local move). This comparison shows how landscape structure directly determines search difficulty, even in simple 1D cases.

\subsection{2D Landscapes}

%
\renewcommand{\mysizeh}{0.2in}
\newcommand{\mysizev}{0.1in}
\begin{figure*}[p]
\vspace{-0.2in}
\centering%
\includegraphics[width=.29\columnwidth]{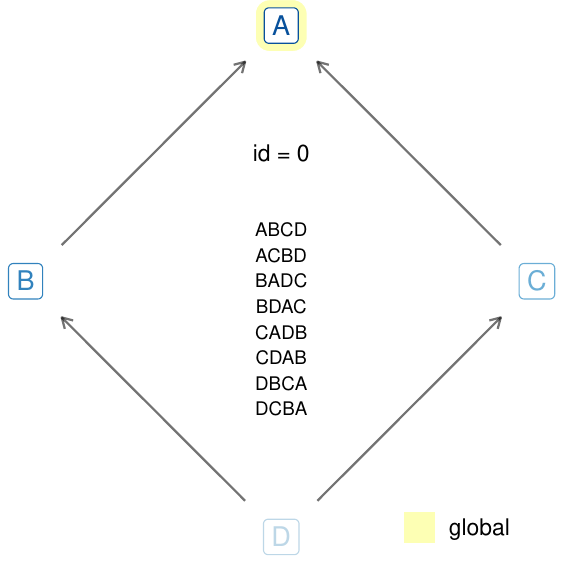}%
\hspace{\mysizeh}%
\includegraphics[width=.29\columnwidth]{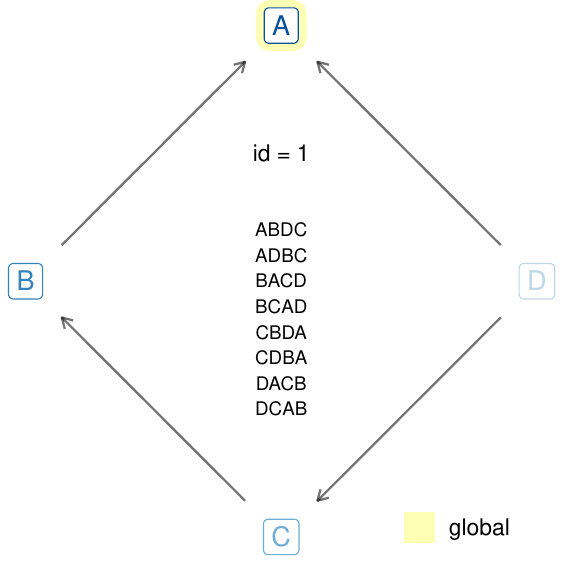}%
\hspace{\mysizeh}%
\includegraphics[width=.29\columnwidth]{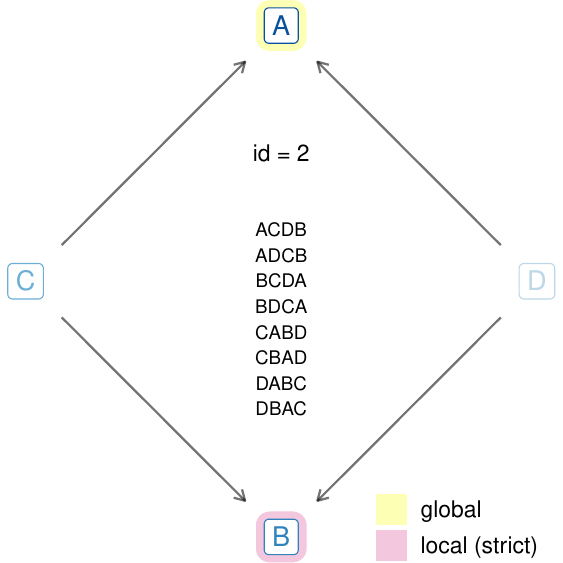}%
\vspace{\mysizev}\\
\includegraphics[width=.29\columnwidth]{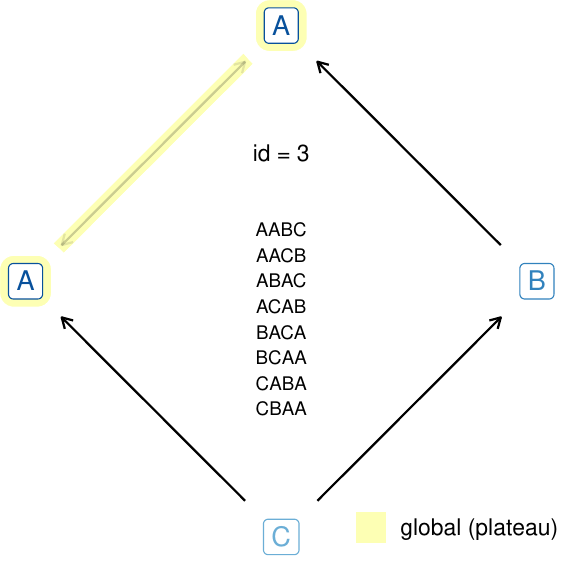}%
\hspace{\mysizeh}%
\includegraphics[width=.29\columnwidth]{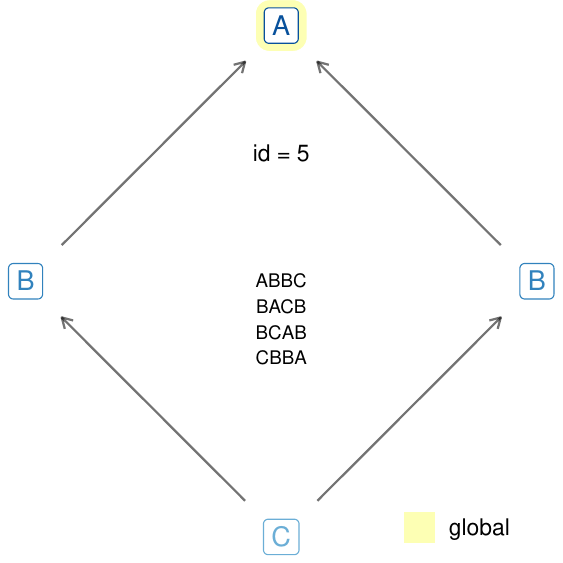}%
\hspace{\mysizeh}%
\includegraphics[width=.29\columnwidth]{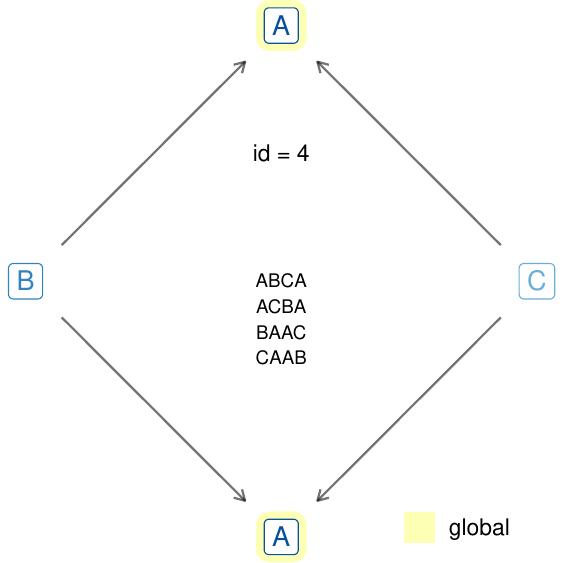}%
\vspace{\mysizev}\\
\includegraphics[width=.29\columnwidth]{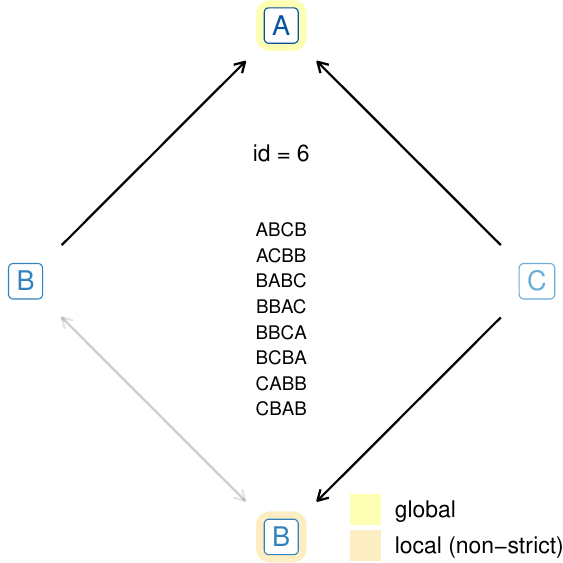}%
\hspace{\mysizeh}%
\includegraphics[width=.29\columnwidth]{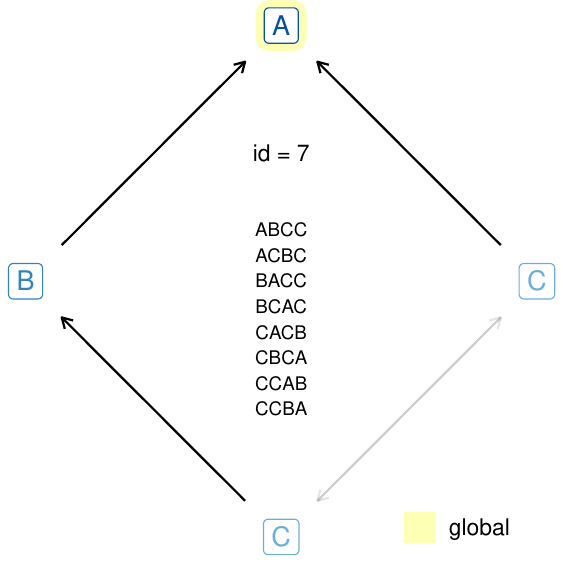}%
\hspace{\mysizeh}%
\includegraphics[width=.29\columnwidth]{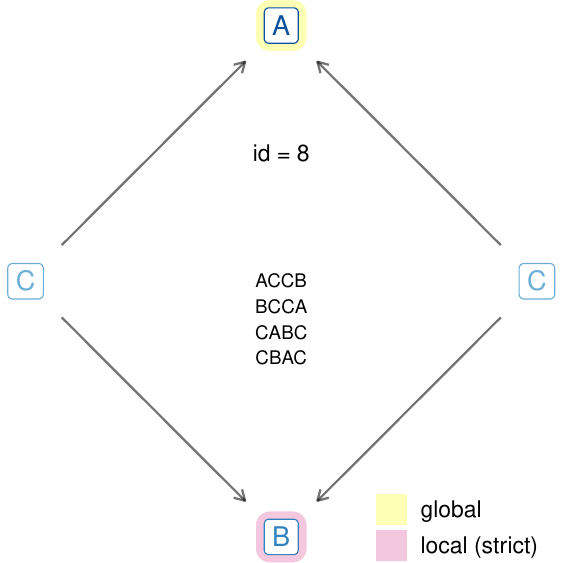}%
\vspace{\mysizev}\\
\includegraphics[width=.29\columnwidth]{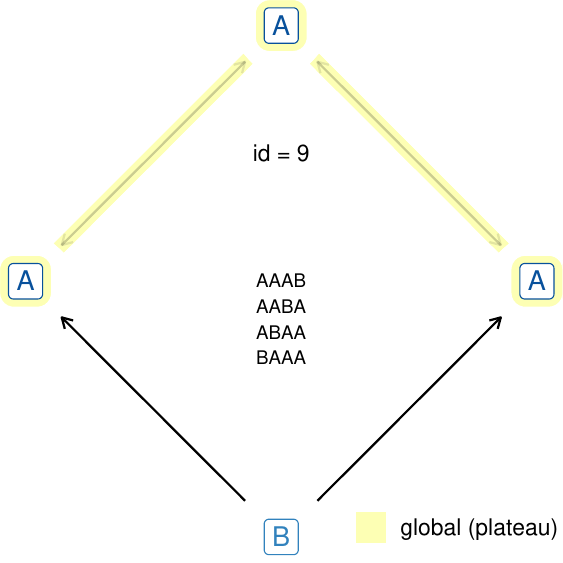}%
\hspace{\mysizeh}%
\includegraphics[width=.29\columnwidth]{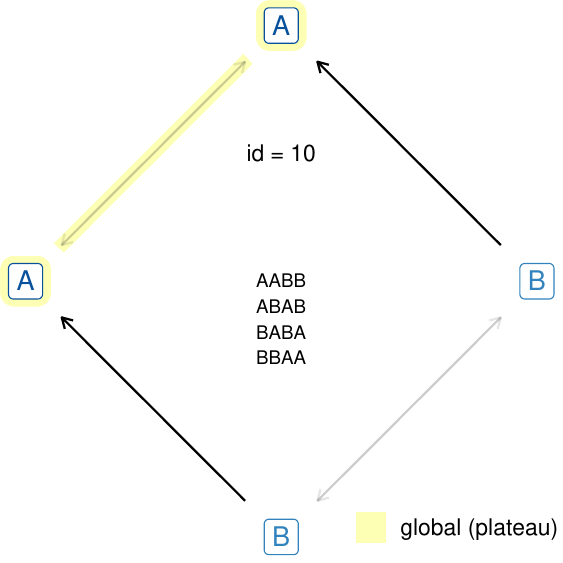}%
\hspace{\mysizeh}%
\includegraphics[width=.29\columnwidth]{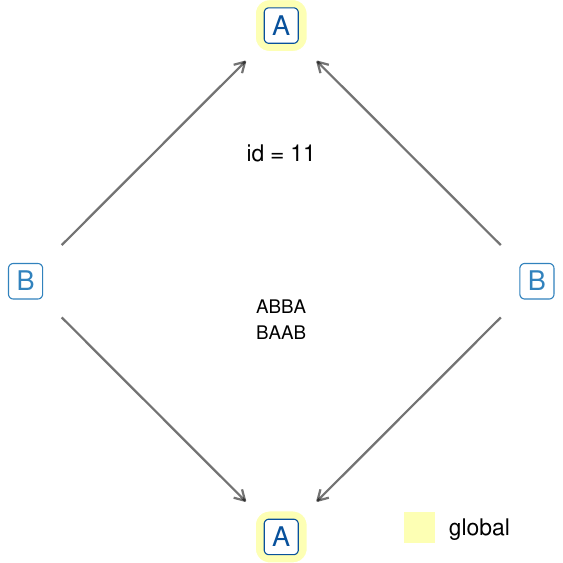}%
\vspace{\mysizev}\\
\includegraphics[width=.29\columnwidth]{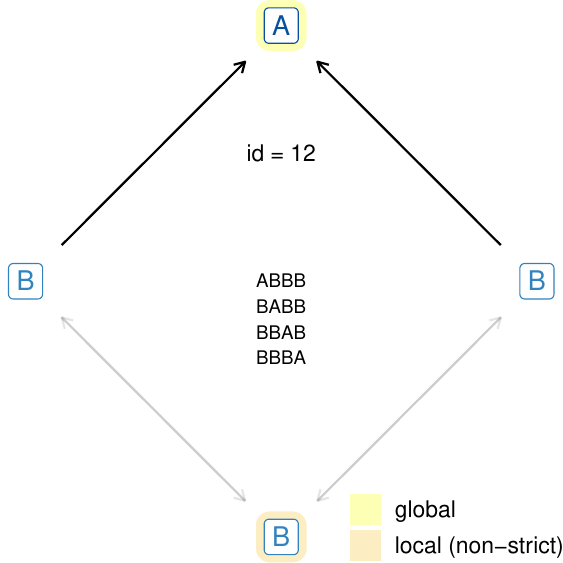}%
\hspace{\mysizeh}%
\includegraphics[width=.29\columnwidth]{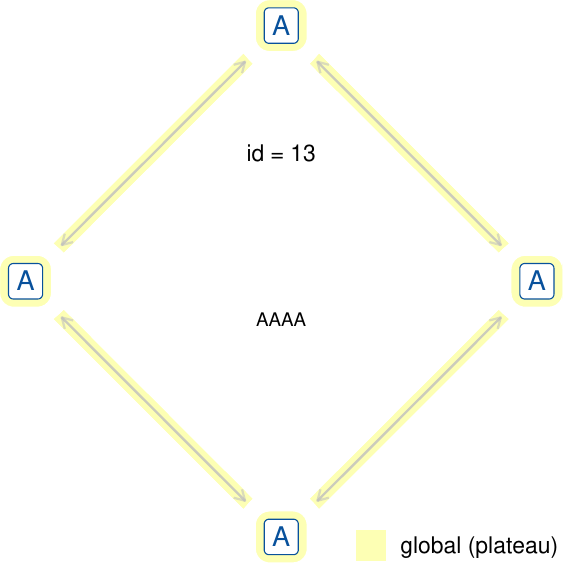}%
\caption{The 14 two-dimensional invariant landscape classes.}
\label{fig:landscapes2}
\end{figure*}

We now present the inventory of 2D invariant landscape classes. As pointed out in Section~\ref{sec:count_rank_invariant_functions}, rank-invariant function classes for $n=2$ comprise $75$ distinct configurations (Table~\ref{tab:partitions_n2}). Through systematic enumeration of all possible automorphisms, we identified \textbf{$\mathbf{14}$ two-dimensional invariant landscape classes}, presented in Figure~\ref{fig:landscapes2}. Solution ranks are denoted by the sequence of rank symbols. For example, for ID \texttt{1}, the ranking on top is $(r_f(00), r_f(01), r_f(10), r_f(11))=$ \A\B\D\C. This classification reveals fundamental structural properties that persist under translation and rotation operations, providing a reduced yet complete set of 2D landscapes.

\paragraph{Classes.}%
The inventory reveals a pronounced contrast: injective functions collapse into just $3$ different invariant classes---reflecting their high symmetry---while the $11$ other classes emerge from functions with $3$ or fewer ranks. This disparity shows how equivalent ranks generate a wider variety of topological configurations.
The automorphisms further emphasize this distinction. Six classes---including all injective functions and some with $3$ ranks---exhibit the complete set of $8$ translation and rotation symmetries. Another six classes only have $4$ symmetries, while the remaining two have just $2$ and $1$ symmetry, respectively. This reduction occurs because certain automorphisms map the landscape into itself due to rank equivalence.

A key distinction from the one-dimensional case is the emergence of \emph{deceptive} landscapes~\citep{Goldberg89}.
A deceptive landscape contains suboptimal local optima (traps), misleading search algorithms away from the global optimum.
In Figure~\ref{fig:landscapes2}, strict local suboptima---solutions that outperform all their neighbors yet remain globally suboptimal---are highlighted in pink, while non-strict suboptima---those with at least one equivalent neighbor---appear in orange. This observation demonstrates how the structural differences between 1D and 2D landscapes introduce an expanding spectrum of challenges for search optimization. Of particular note is the absence of suboptimal plateaus, a characteristic that will emerge in 3D landscapes.

\newcommand{\cmark}{$\ovoid$}%
\newcommand{\xmark}{\ding{53}}%
\begin{table}[t!]
\caption{Properties of the 14 two-dimensional invariant landscape classes.
\smallskip}
\label{tab:properties_n2}
\centering%
\begin{tabular}{r|cc|ccc|l}
\toprule
\textbf{id} & \textbf{ranks} & \textbf{rankings} & \textbf{deceptive} & \textbf{neutral} & \textbf{plateau} & \textbf{comments} \\
\midrule
0  & 4     & 8          & \xmark     & \xmark  & \xmark & {\small Injective} \\
1  & 4     & 8          & \xmark     & \xmark  & \xmark & {\small Injective} \\
2  & 4     & 8          & \cmark     & \xmark  & \xmark & {\small Injective, Min. deceptive} \\
3  & 3     & 8          & \xmark     & \cmark  & \cmark & \\
4  & 3     & 4          & \xmark     & \xmark  & \xmark & \\
5  & 3     & 4          & \xmark     & \xmark  & \xmark & OneMax \\ 
6  & 3     & 8          & $\triangle$   & \cmark  & \xmark & \\
7  & 3     & 8          & \xmark     & \cmark  & \xmark & \\
8  & 3     & 4          & \cmark     & \xmark  & \xmark & \\
9  & 2     & 4          & \xmark     & \cmark  & \cmark & {\small Boolean, SAT-like} \\
10 & 2     & 4          & \xmark     & \cmark  & \cmark & {\small Boolean, Walsh order 1} \\
11 & 2     & 2          & \xmark     & \xmark  & \xmark & {\small Boolean, Walsh order 2} \\
12 & 2     & 4          & $\triangle$   & \cmark  & \xmark & {\small Boolean} \\
13 & 1     & 1          & \xmark     & \cmark  & \cmark & {\small Boolean} \\
\midrule
\textbf{count}	&	--					&	\textbf{75}					&	\textbf{2$\sim$4}					&	\textbf{7}   &   \textbf{4}	&   --				\\
\bottomrule
\end{tabular}
\end{table}

\paragraph{Properties.}%
Table~\ref{tab:properties_n2} summarizes the key characteristics of the $14$ invariant classes of 2D landscapes, including deceptiveness. Further details are provided in Appendix (Table~\ref{tab:neutrality_n2})
and as supplementary material at \url{https://doi.org/10.5281/zenodo.18492019}. It reveals several critical insights into the topological features of these landscapes.
Firstly, two to four landscape classes are deceptive, with ID \texttt{2} representing the sole injective case where deceptiveness occurs---this is the famous minimal deceptive problem from \citet{goldberg1987simple}. The $\triangle$ symbol in landscapes with IDs \texttt{6} and \texttt{12} indicates the presence of non-strict suboptima, which are equivalent to at least one of their neighbors. Such configurations may mislead search algorithms, though their impact can be mitigated compared to strict local optimality.

Secondly, half of the landscape classes exhibit neutrality, with four also containing plateaus. A landscape is \emph{neutral} if it includes at least one neutral edge---a pair of neighbors with equivalent fitness~\citep{ReiSta01}. A landscape has a \emph{plateau} if it contains a connected set of local optima with equal fitness (same rank), where no solution has an improving neighbor~\citep{HooStu05sls-mk}. We exclude degenerate plateaus consisting of a single solution and use ``plateau'' throughout for brevity. Note that all landscapes with plateaus are inherently neutral, since plateaus require at least one neutral edge. However, neutrality does not necessarily imply plateaus, as neutral edges may connect solutions that have improving neighbors.
Such a neutrality naturally occurs in several functions with one to three ranks. While plateaus typically present stagnation risks for local search algorithms by trapping them in flat regions, this concern is mitigated in these 2D cases since all plateaus consist exclusively of global optima. Nonetheless, search methods must still recognize when they have reached such plateaus to avoid premature termination or unnecessary exploration.
Low-rank landscapes (IDs \texttt{9}–\texttt{13}) are analogous to Boolean functions, with ID \texttt{9} mirroring a \mbox{$2$-SAT} clause.
These low-rank landscapes 
typically feature large plateaus 
which may either hinder or facilitate exploration, depending on the algorithm's ability to exploit neutrality.
Landscapes with IDs \texttt{10} and \texttt{11} are associated with Walsh functions of order~1 and~2, respectively~\citep{goldberg1989genetic}. Walsh functions form a basis for the space of pseudo-Boolean functions~\citep{odonnell2014analysis}. The order-1 Walsh functions involve independent binary variables, whereas the order-2 Walsh functions depend on the interaction of both variables. The inventory shows that they fall into different invariant landscape classes.%

Among the 14 invariant classes, five (IDs \texttt{0}, \texttt{1}, \texttt{4}, \texttt{5}, and \texttt{11}) stand out as structurally simple: they exhibit neither deceptiveness nor neutrality, making them relatively simple for search. Notably, ID \texttt{5} corresponds to the well-known OneMax problem class~\citep{DroJanWeg06,Jansen13}---generalizing basic OneMax to minimizing Hamming distance to any target bitstring. Interestingly, OneMax (ID \texttt{5}) is not necessarily the simplest landscape for local search, as IDs \texttt{4} and \texttt{11} contain multiple global optima. This distinction highlights how even ``simple'' landscapes can present nuanced challenges depending on their structure. None of the remaining landscape classes combine (strong) deceptiveness with neutrality---a combination that, as we will demonstrate, emerges in 3D landscapes.%

\paragraph{Performance.}%
We now examine the expected performance of two baseline algorithms across these $14$ landscape classes. We consider landscape-invariant local search strategies that serve as building blocks for many sophisticated algorithms: a best-improvement hill climber, which selects the best strictly improving neighbor at each iteration, and a first-improvement hill climber, which selects the first strictly improving neighbor encountered during random exploration of the neighborhood (without replacement). Both algorithms terminate when there is no strictly improving neighbor. For each approach, we first compute the exact success rate, that is the probability of reaching a global optimum from an arbitrary starting solution. In addition, we measure their expected runtime (ERT) in function evaluations~\citep{HooStu05sls-mk,AugHan2005lrcmaes} by restarting the search from scratch whenever it gets stuck in a suboptimum, repeating until a global optimum is found. Results are provided in Table~\ref{tab:perf_n2}, with more details in Appendix (Tables~\ref{tab:bihc_n2} and~\ref{tab:fihc_n2}) and
supplementary material available at \url{https://doi.org/10.5281/zenodo.18492019}.

\begin{table}[t!]
\caption{Expected performance of best- and first-improvement hill climber (single- and multi-start) on the 14 two-dimensional invariant landscape classes.
ERT stands for expected runtime.
See Tables~\ref{tab:bihc_n2}--\ref{tab:fihc_n2} in Appendix for details.
\smallskip}
\label{tab:perf_n2}
\centering%
\begin{tabular}{r|rr|rr}
\toprule
 & \multicolumn{2}{c|}{\textbf{best-improvement}}
 & \multicolumn{2}{c}{\textbf{first-improvement}} \\
\textbf{id} & \textbf{success rate} & \textbf{multi-start ERT} & \textbf{success rate} & \textbf{multi-start ERT} \\
\midrule
0  & 1.000 & 5.000 & 1.000 & 4.375 \\
1  & 1.000 & 5.000 & 1.000 & 4.750 \\
2  & 0.750 & 5.333 & 0.500 & 7.000 \\
3  & 1.000 & 4.000 & 1.000 & 3.812 \\
4  & 1.000 & 4.000 & 1.000 & 3.500 \\
5  & 1.000 & 5.000 & 1.000 & 4.375 \\
6  & 0.750 & 5.333 & 0.625 & 5.800 \\
7  & 1.000 & 5.000 & 1.000 & 4.500 \\
8  & 0.750 & 5.333 & 0.500 & 7.000 \\
9  & 1.000 & 3.500 & 1.000 & 3.250 \\
10 & 1.000 & 4.000 & 1.000 & 3.750 \\
11 & 1.000 & 4.000 & 1.000 & 3.500 \\
12 & 0.750 & 5.333 & 0.750 & 5.000 \\
13 & 1.000 & 3.000 & 1.000 & 3.000 \\
\bottomrule
\end{tabular}
\end{table}

We observe that the best-improvement hill climber consistently achieves a success rate equal to or better than the first-improvement variant across all 14 landscape classes. However, this advantage comes at a computational cost: when considering expected runtime, the multi-start first-improvement hill climber outperforms best-improvement in $10$ out of $14$ classes, while being slower in only $3$. This trade-off highlights the practical efficiency of first-improvement despite its lower success probability per restart. This is of particular interest when deploying local search to solve small sub-problems as a subroutine for larger-scale problems.
Interestingly, the multi-start best-improvement hill climber retains a significant advantage in $3$ landscape classes, all of which belong to the $4$ classes that suffer from deceptiveness.

\bigskip\noindent%
The diversity of landscape features---from deceptiveness to neutrality and plateaus---reveals the spectrum of challenges present even in simple 2D landscapes. These challenges highlight the need for advanced search strategies. While algorithms tailored for neutral landscapes may struggle on deceptive ones, alternative approaches may better navigate these complexities. This foundation sets the stage for exploring higher-dimensional landscapes, where the interplay between deceptiveness, neutrality, and plateaus is expected to intensify.

\subsection{3D Landscapes}

Extending our methodology to 3D landscapes, we enumerate the invariant classes derived from the $545\,835$ rank-invariant functions (Section~\ref{sec:count_rank_invariant_functions}, Table~\ref{tab:partitions}), yielding a total of 
\textbf{$\mathbf{11\,991}$ three-dimensional invariant landscape classes}.
Given the extensive size of this inventory, we have made the complete collection available as supplementary material at \url{https://doi.org/10.5281/zenodo.18492019}. 
In this section, we present selected representative examples along with summary statistics of their structural properties, highlighting the increased complexity and topological diversity that emerges in higher dimensions.

\paragraph{Examples.}%
Figure~\ref{fig:landscapes3} shows $15$ representative examples out of the $11\,991$ distinct 3D invariant landscape classes. We highlight that IDs \texttt{1371} and \texttt{1580} correspond to the landscapes resulting from the example functions elaborated in Section~\ref{sec:summary} (Figure~\ref{fig:summary_example}). Even with this limited selection of $15$, the diversity of landscapes is evident in terms of the number of ranks, global optima, local suboptima (strict or non-strict), neutrality, and plateaus (optimal or suboptimal).

%
\renewcommand{\mysizeh}{0.1in}
\renewcommand{\mysizev}{0.2in}
\begin{figure*}[p]
\centering%
\includegraphics[width=.32\columnwidth]{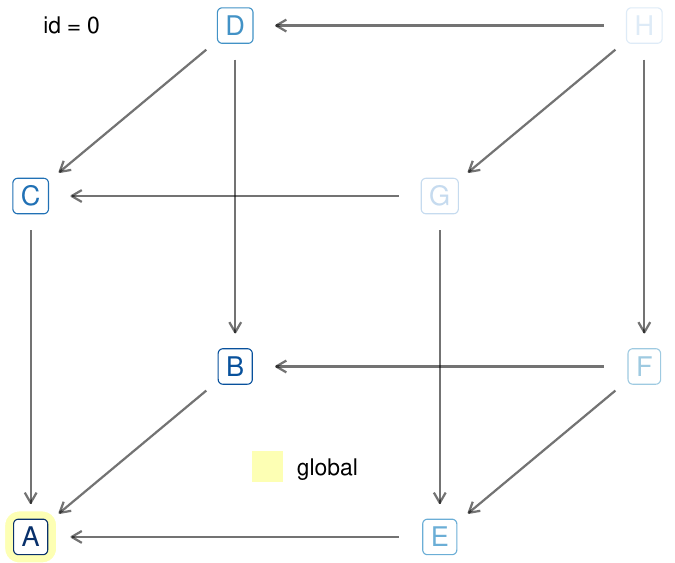}%
\hspace{\mysizeh}%
\includegraphics[width=.32\columnwidth]{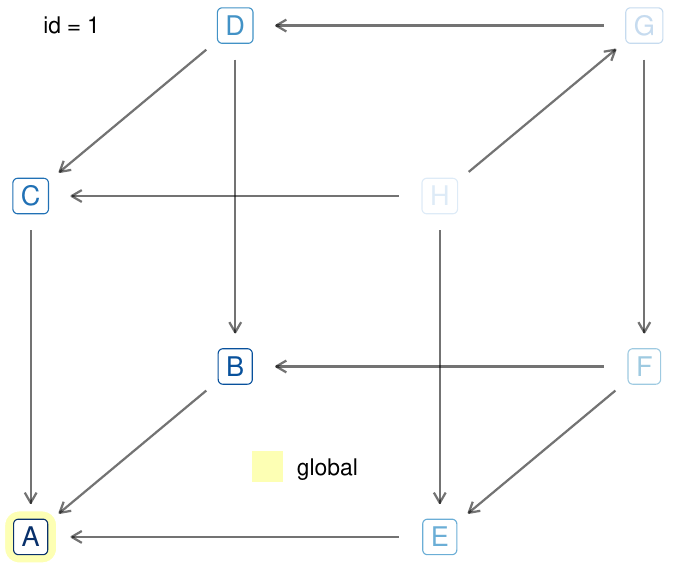}%
\hspace{\mysizeh}%
\includegraphics[width=.32\columnwidth]{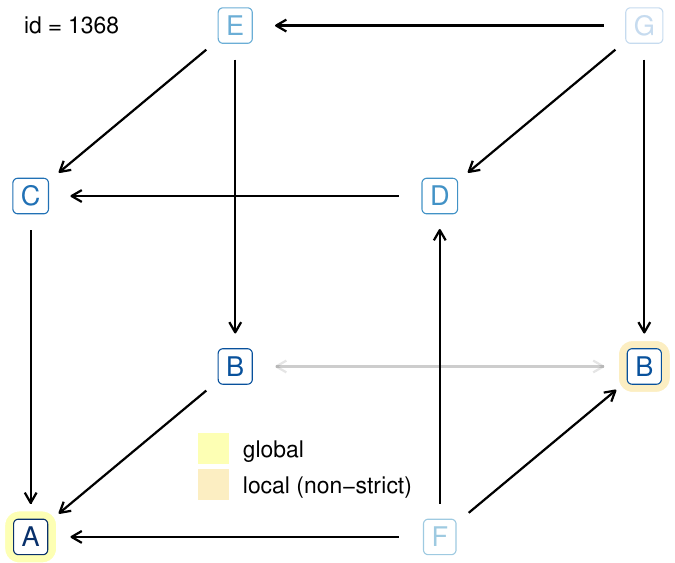}%
\vspace{\mysizev}\\
\includegraphics[width=.32\columnwidth]{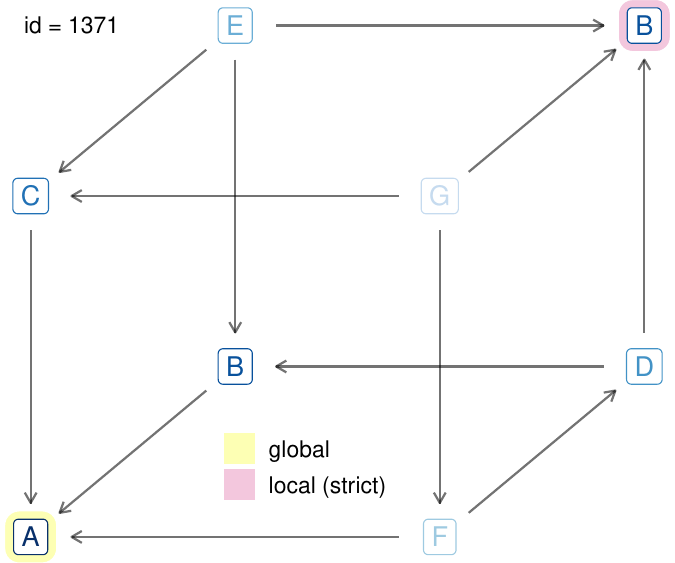}%
\hspace{\mysizeh}%
\includegraphics[width=.32\columnwidth]{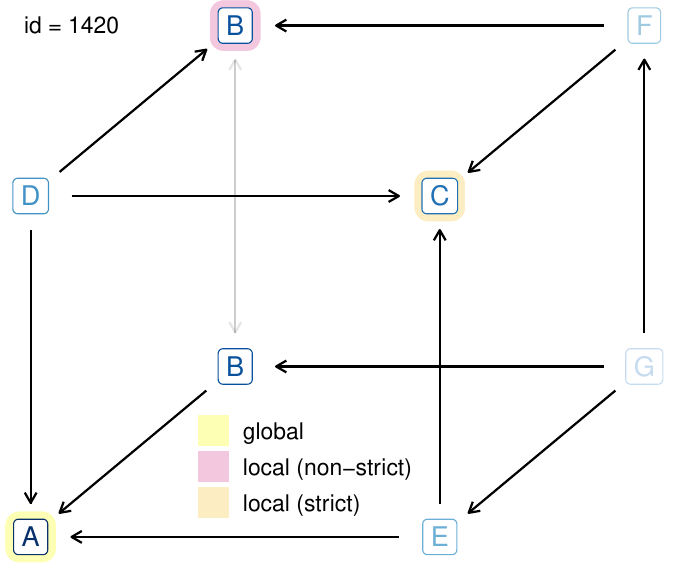}%
\hspace{\mysizeh}%
\includegraphics[width=.32\columnwidth]{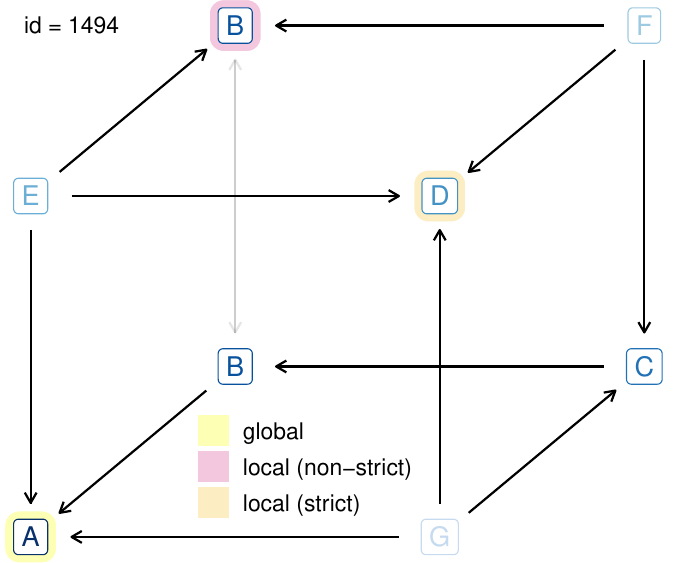}%
\vspace{\mysizev}\\
\includegraphics[width=.32\columnwidth]{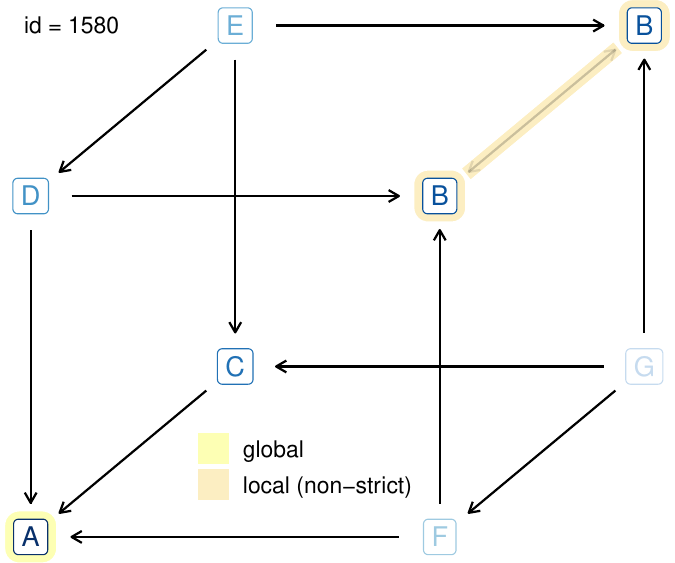}%
\hspace{\mysizeh}%
\includegraphics[width=.32\columnwidth]{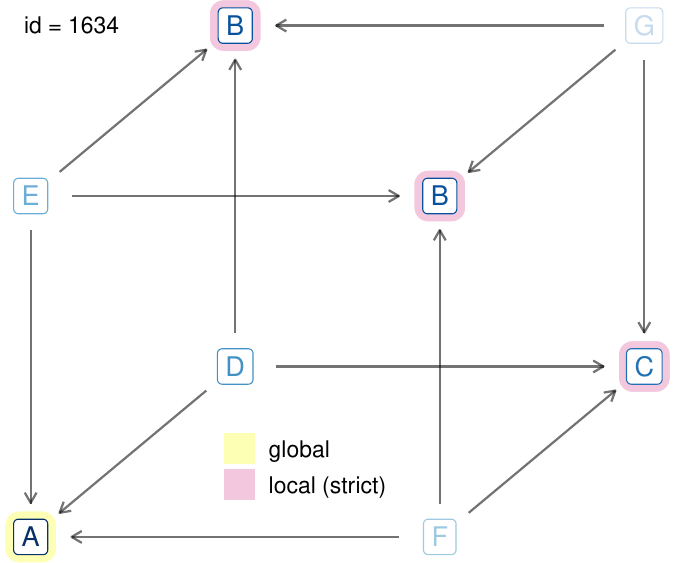}%
\hspace{\mysizeh}%
\includegraphics[width=.32\columnwidth]{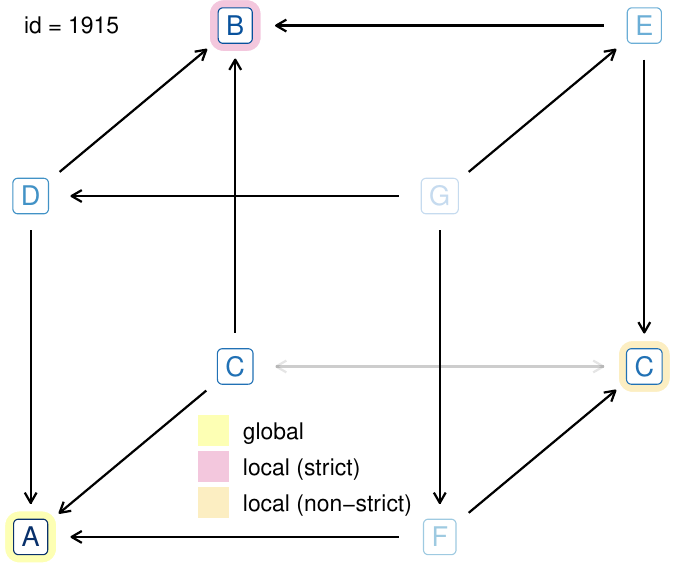}%
\vspace{\mysizev}\\
\includegraphics[width=.32\columnwidth]{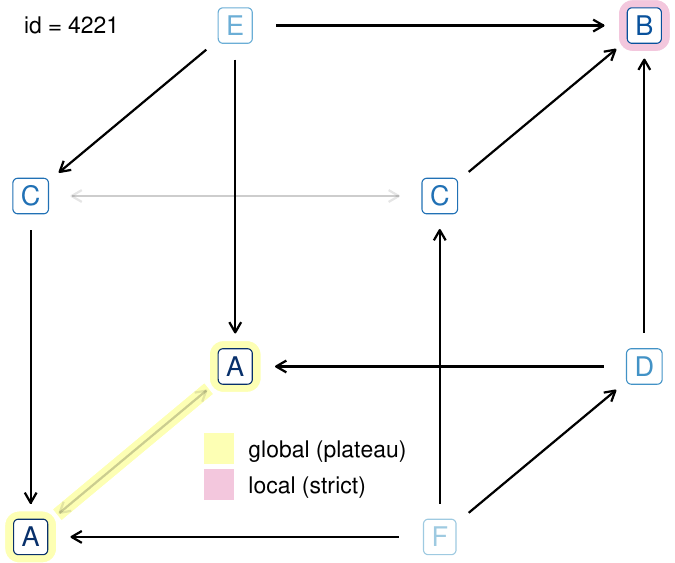}%
\hspace{\mysizeh}%
\includegraphics[width=.32\columnwidth]{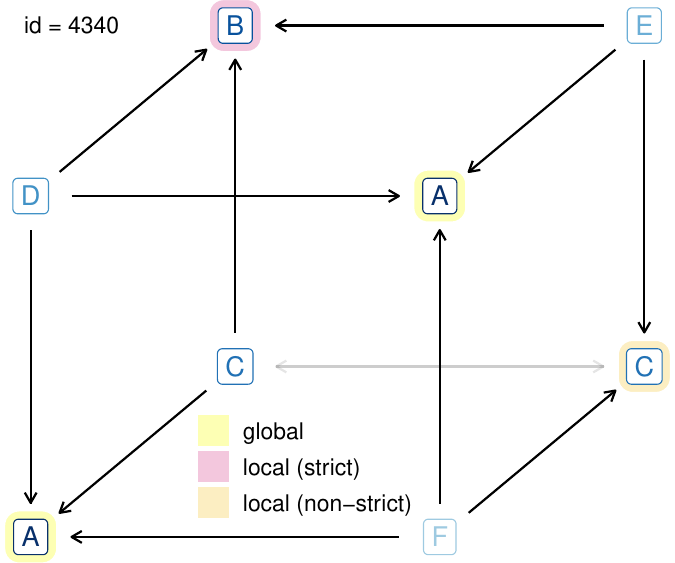}%
\hspace{\mysizeh}%
\includegraphics[width=.32\columnwidth]{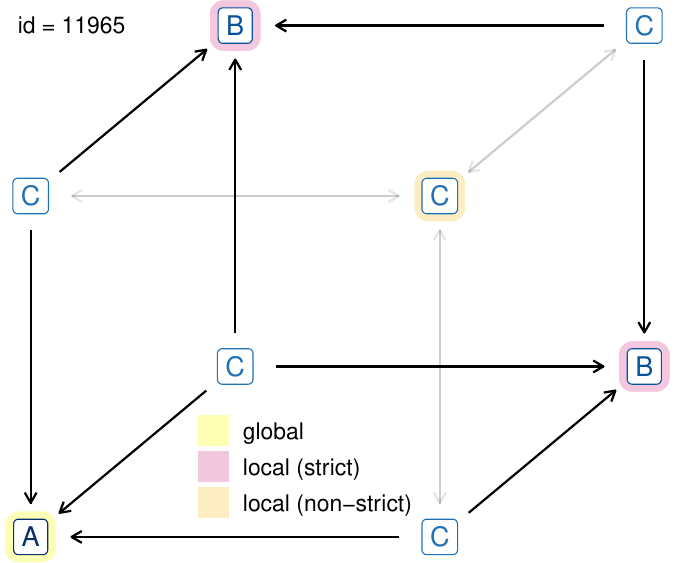}%
\vspace{\mysizev}\\
\includegraphics[width=.32\columnwidth]{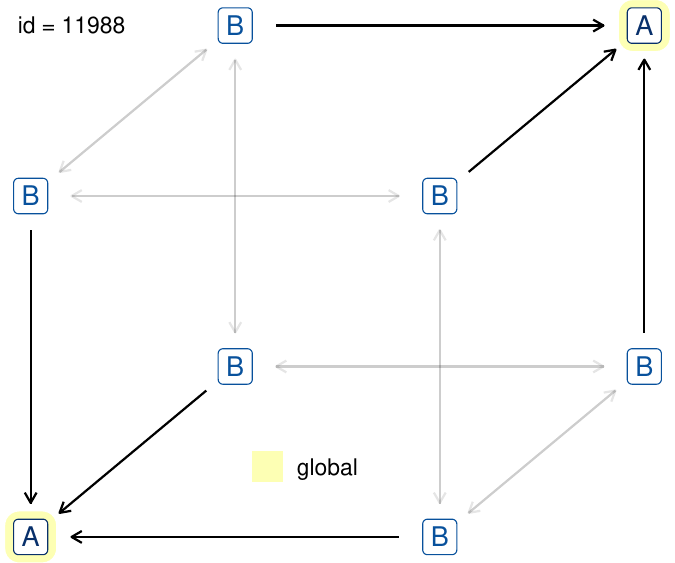}%
\hspace{\mysizeh}%
\includegraphics[width=.32\columnwidth]{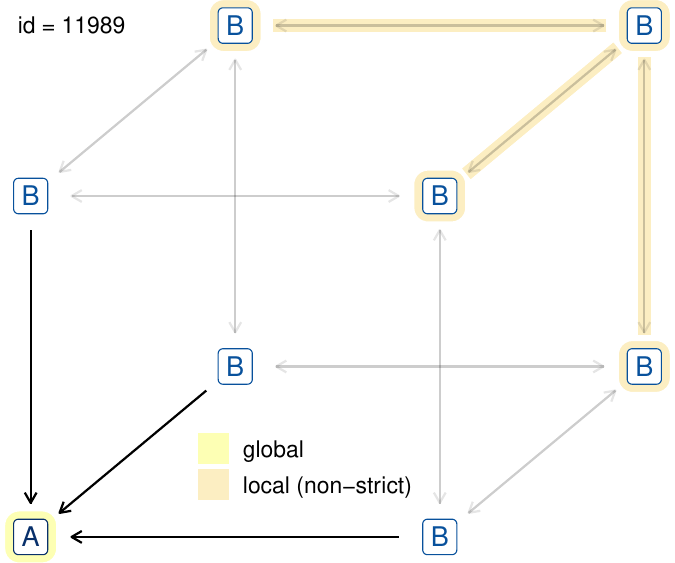}%
\hspace{\mysizeh}%
\includegraphics[width=.32\columnwidth]{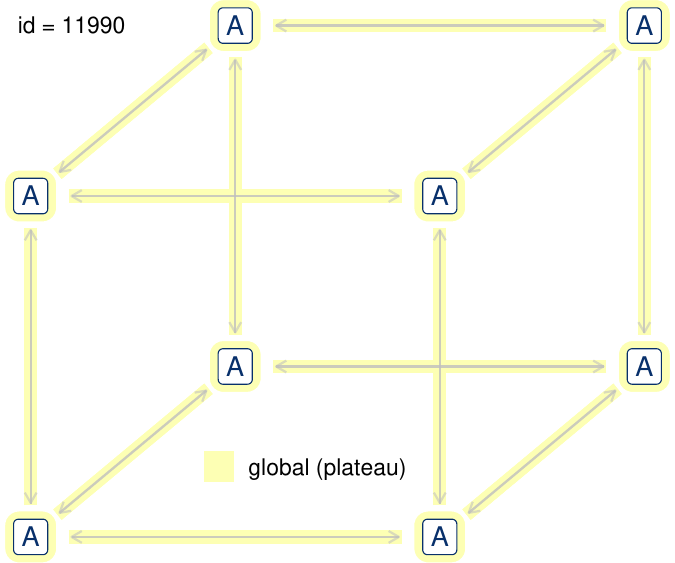}%
\caption{Examples from the 11\,991 three-dimensional invariant landscape classes.} 
\label{fig:landscapes3}
\end{figure*}

For instance, while landscapes ID \texttt{11988} and ID \texttt{11989} appear similar at first glance, they present distinct challenges: ID \texttt{11989} (which corresponds to a 3-SAT clause) features a suboptimal plateau that is absent in ID \texttt{11988}. This subtle difference highlights the nuanced difficulties that can emerge even among visually comparable landscapes. Furthermore, the 3D inventory includes landscapes that combine strong deceptiveness with neutrality (e.g., IDs \texttt{1494}, \texttt{1915}, \texttt{4221}), or even with a suboptimal plateau (e.g., ID \texttt{1580})---a combination not observed in smaller landscapes with $n < 3$.%

Given the large number of classes, characterizing each one exhaustively is infeasible. Instead, we summarize their properties below to provide a comprehensive overview of their structural diversity.

\paragraph{Properties.}%

\begin{figure*}[!t]
\centering%
\includegraphics[width=.49\columnwidth]{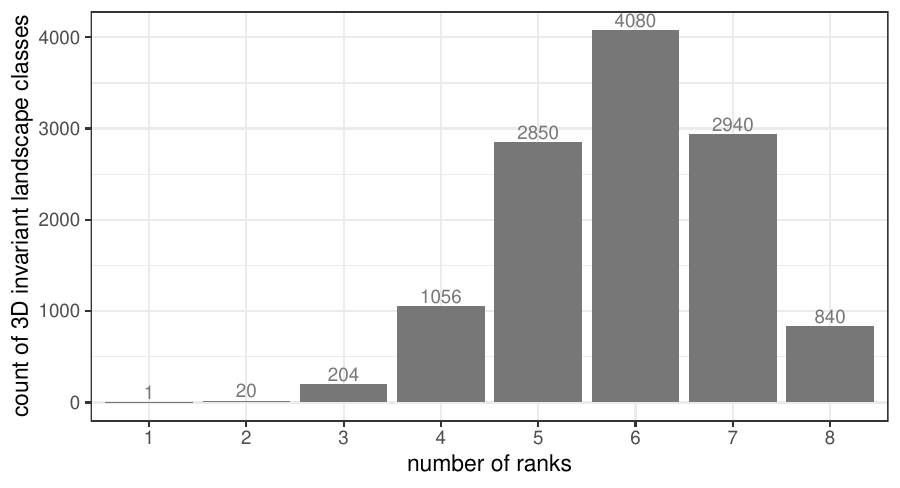}\hfill%
\includegraphics[width=.49\columnwidth]{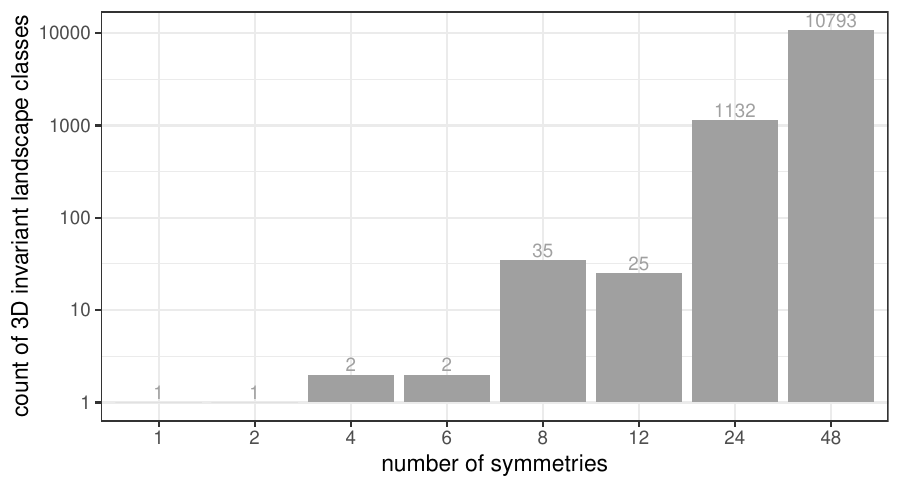}\\
\includegraphics[width=.49\columnwidth]{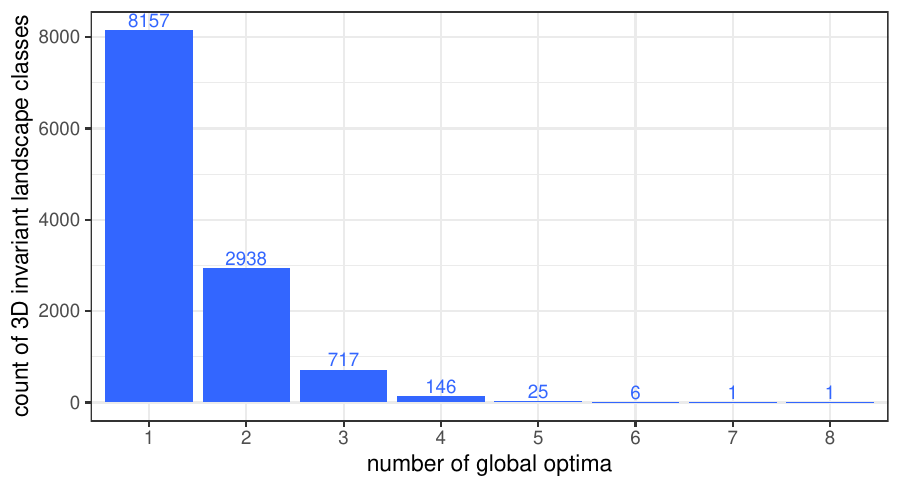}\hfill%
\includegraphics[width=.49\columnwidth]{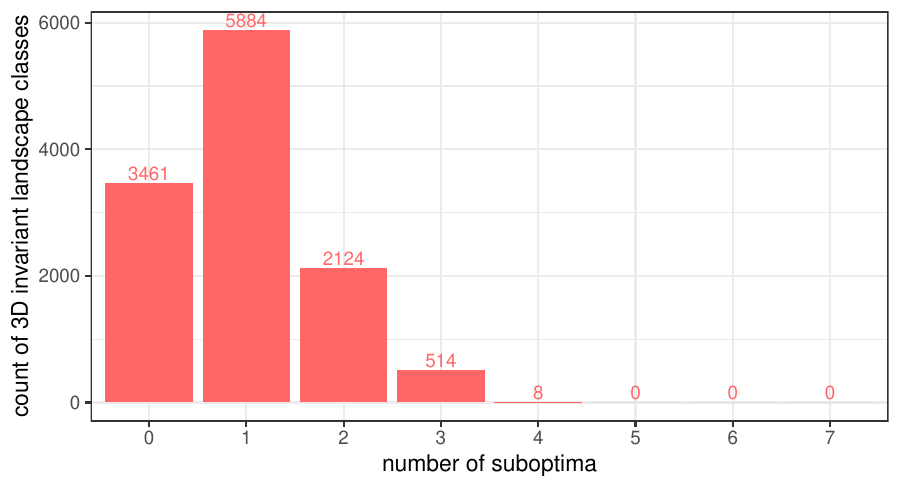}\\
\includegraphics[width=.49\columnwidth]{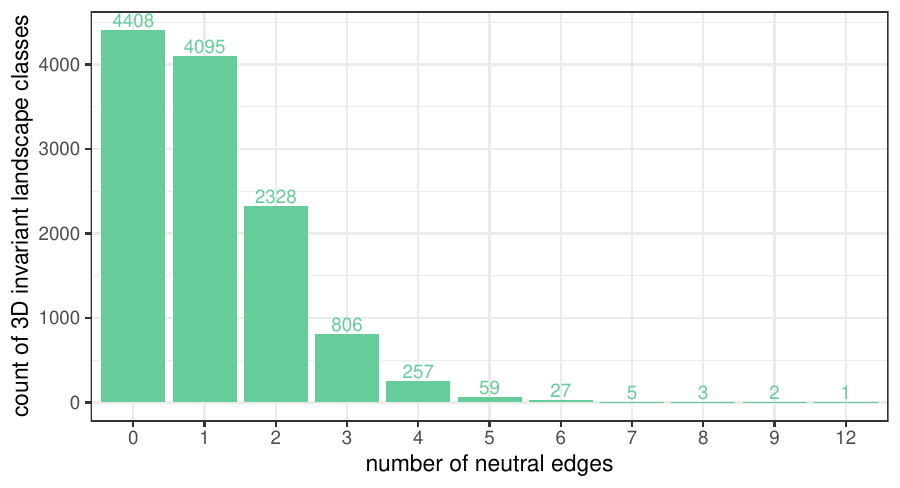}\hfill
\includegraphics[width=.49\columnwidth]{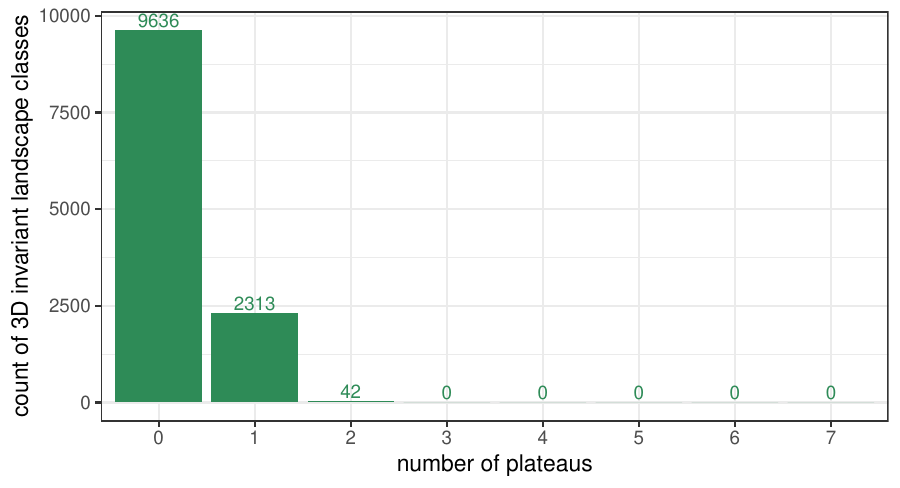}\\
\hfill%
\includegraphics[width=.49\columnwidth]{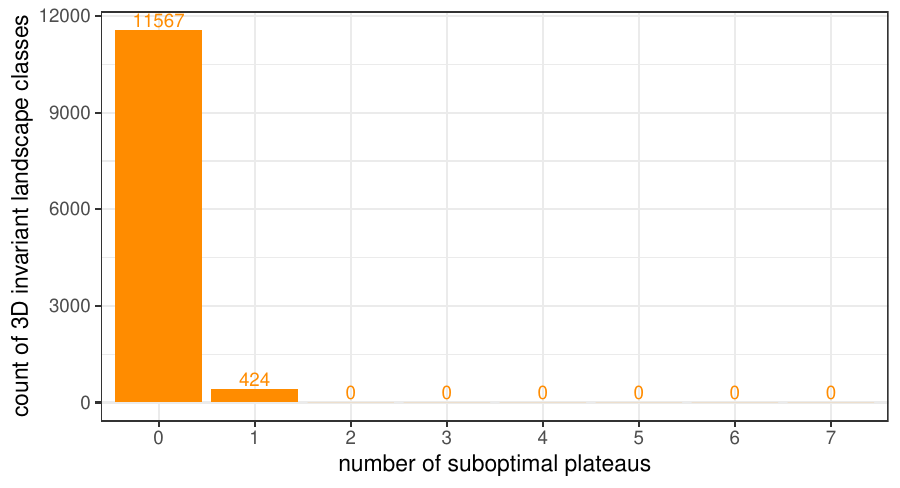}\\
\caption{Distribution of the three-dimensional invariant landscape classes by key properties: number of ranks, symmetries, global optima, suboptima, neutral edges, plateaus, and suboptimal plateaus.}
\label{fig:features_n3}
\end{figure*}

Figure~\ref{fig:features_n3} summarizes the distribution of 3D landscape classes according to key properties: number of ranks, symmetries, global optima, local suboptima, neutral edges, and plateaus (all or suboptimal only). Notably, injective functions collapse into just $840$ invariant classes ($7\%$), while the remaining $93\%$ emerge from functions with seven or fewer ranks. Over $90\%$ of classes exhibit the full set of $48$ symmetries, with most remaining classes having $24$ symmetries---except for a handful of exceptions.

About $30\%$ of classes contain multiple global optima (connected or disconnected), while over $70\%$ are deceptive, featuring one or more local suboptima. Neutrality is also prevalent, with over $63\%$ of classes containing at least one neutral edge---a pair of neighbors with the same rank. This results in one or two plateaus in nearly $20\%$ of classes and a suboptimal plateau in $3.5\%$ of them.

Table~\ref{tab:properties_n3} synthesizes these properties by partitioning classes according to the presence or absence of deceptiveness, neutrality, and plateaus. The most common category---nearly $35\%$ of all classes---combines neutrality and deceptiveness without plateaus. The second most frequent category (over $25\%$) includes deceptive landscape classes without neutrality. The remaining classes are roughly evenly distributed (around $10\%$ each) among those that are neither deceptive nor neutral nor contain plateaus, those with only neutrality, those with neutrality and plateaus, and those combining all three properties. This distribution highlights the significant diversity of classes across these landscape  properties---a diversity whose implications for algorithm performance remain to be explored.

\begin{table}[t!]
\caption{Number and percentage of the three-dimensional invariant landscape classes by combination of properties (deceptiveness, neutrality, and plateaus).\smallskip}%
\label{tab:properties_n3}
\centering%
\begin{tabular}{ccc|rr}
\toprule
\textbf{deceptive} & \textbf{neutral} & \textbf{plateau}~ & ~ \textbf{count} & \textbf{percentage} \\
\midrule
\xmark     & \xmark     & \xmark     & $1\,233$ &   $10.28\%$   \\
\midrule
\cmark     & \xmark     & \xmark     & $3\,175$ &   $26.48\%$   \\
\xmark     & \cmark     & \xmark     & $1\,098$ &   $9.16\%$    \\
\midrule
\cmark     & \cmark     & \xmark     & $4\,130$ &   $34.44\%$   \\
\xmark     & \cmark     & \cmark     & $1\,130$ &   $9.42\%$    \\
\midrule
\cmark     & \cmark     & \cmark     & $1\,225$ &   $10.22\%$   \\

\bottomrule
& & \textbf{total}	& 11\,991 & $100\%$ \\
\end{tabular}
\end{table}

\paragraph{Performance.}%

\begin{figure*}[t!]
\centering%
\includegraphics[width=.49\columnwidth]{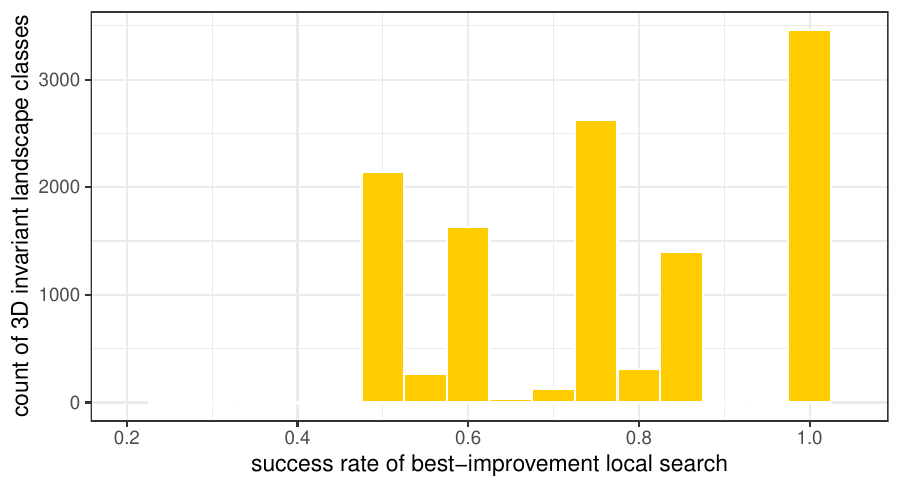}\hfill%
\includegraphics[width=.49\columnwidth]{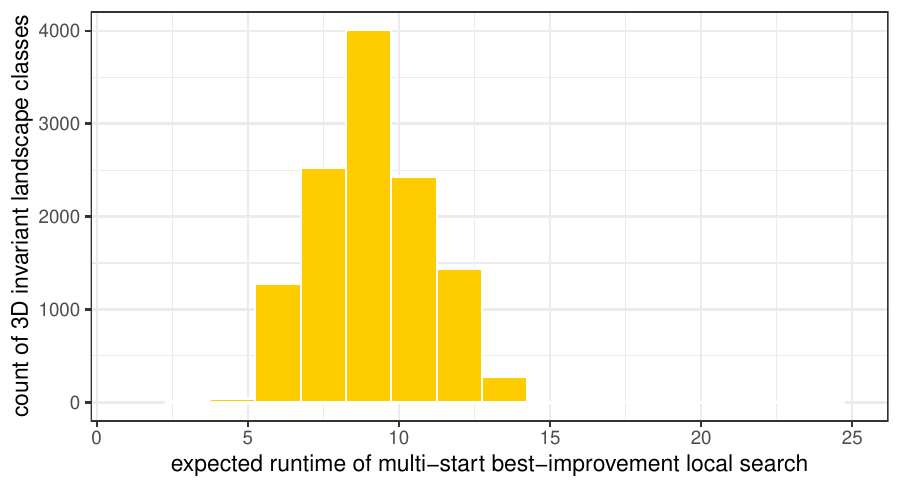}\\
\includegraphics[width=.49\columnwidth]{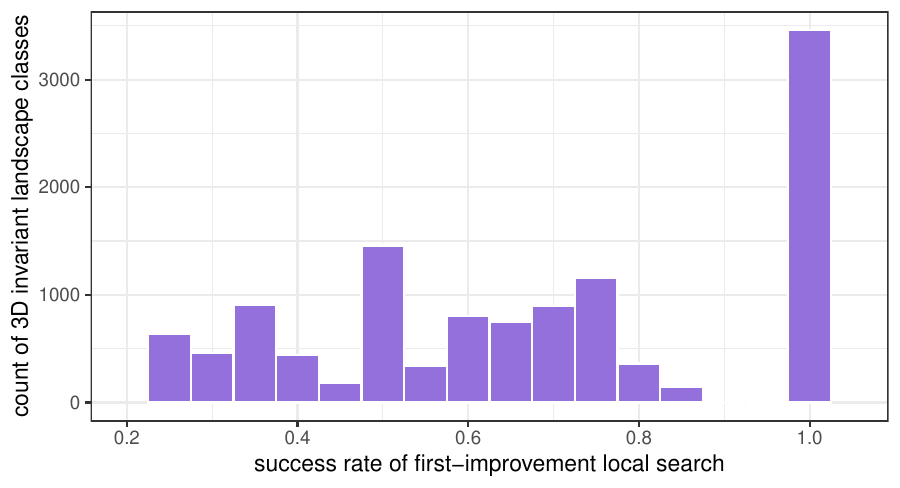}\hfill%
\includegraphics[width=.49\columnwidth]{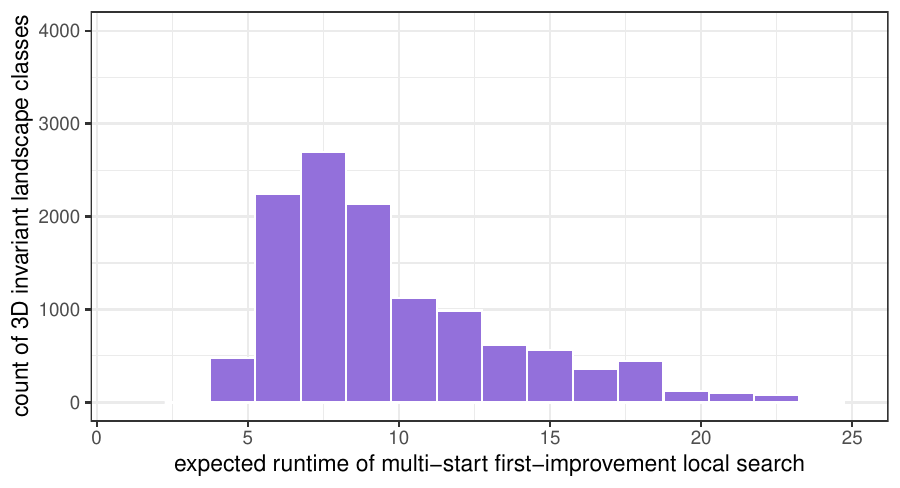}\\
%
\caption{Distribution of the three-dimensional invariant landscape classes by success rate and expected runtime for best-improvement and first-improvement hill climbers.}
\label{fig:perf_n3}
\end{figure*}

We evaluate the performance of best- and first-improvement hill climbers, examining their success rates and expected running time (ERT) in a multi-start setting (Figure~\ref{fig:perf_n3}). The first-improvement variant exhibits a broader range of performance values compared to the best-improvement approach.

Further investigations reveal that the best-improvement hill climber achieves a higher success rate than the first-improvement variant for $7\,268$ classes ($61\%$), while the first-improvement approach performs better for only $653$ classes ($5\%$). For the remaining $4\,070$ classes ($34\%$), both methods achieve identical success rates. In terms of expected runtime, the multi-start first-improvement hill climber is faster on $7\,064$ classes ($59\%$), while the multi-start best-improvement variant is faster on $4\,916$ classes ($41\%$). Both methods perform equally for $11$ classes (slightly above $0\%$).

\begin{figure*}[!h]
\centering%
\includegraphics[width=.49\columnwidth]{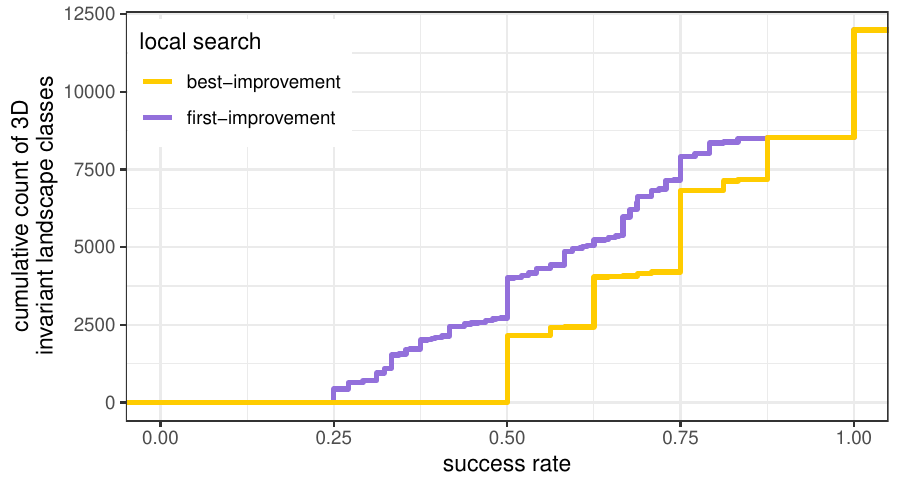}\hfill%
\includegraphics[width=.49\columnwidth]{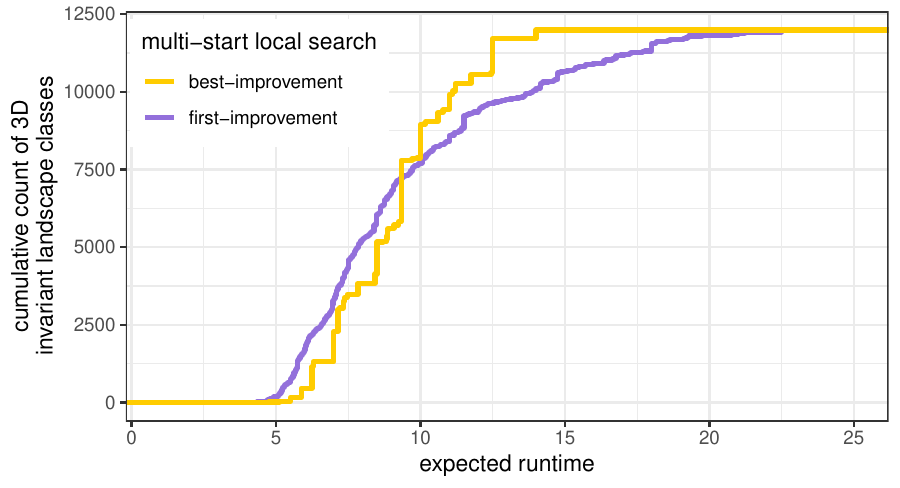}
%
\caption{Cumulative distribution of the three-dimensional invariant landscape classes where the success rate and expected runtime fall below a threshold for best-improvement and first-improvement hill climbers.}
\label{fig:perf_n3_cumulative}
\end{figure*}

To facilitate comparison, Figure~\ref{fig:perf_n3_cumulative} shows the cumulative number of classes for which the success rate (or ERT) falls below a given threshold on the x-axis. The first-improvement hill climber never achieves a better success rate than best-improvement across more classes. However, when random restarts are applied upon stagnation, the first-improvement method solves more classes within a budget of $9$ evaluations or fewer. Conversely, for budgets exceeding $9$, the best-improvement hill climber becomes more effective on more classes. The performance complementarity of these two baseline local search methods further underscores the diversity of 3D invariant landscape classes. This suggests that the 3D inventory already provides a relevant scenario for algorithm recommendation based on landscape properties, even for problems as small as $n=3$ variables.

\bigskip\noindent%
The analysis of 3D invariant landscape classes reveals a rich combination of difficulties---deceptiveness, neutrality, and plateaus---each of which can be toggled on or off (Table~\ref{tab:properties_n3}). At least one example representing every possible combination appears in Figure~\ref{fig:landscapes3}. This inventory provides a compelling argument for focusing on problems as small as dimension $3$. The observed variety not only challenges algorithm performance but also opens avenues for deeper investigation into the interplay between landscape properties and search dynamics.

\section{Concluding Remarks}
\label{sec:conclu}

This study presents an exhaustive inventory of invariant landscape classes for pseudo-Boolean functions of dimensions $n \in \set{1,2,3}$, including non-injective functions. Our analysis identifies $2$ invariant landscape classes for $n=1$, $14$ for $n=2$, and $11\,991$ for $n=3$---totaling $12\,007$ unique classes. This comprehensive catalog is the first systematic enumeration of all possible invariant classes that accounts for both rankings and neighborhood symmetries (translations and rotations). This matters because 
many randomized optimization
algorithms behave identically on problems within the same class, relying solely on relative comparisons and neighborhood relations.

The inventory reveals that non-injective functions dominate, accounting for $78\%$ of classes for $n=2$ and $92\%$ for $n=3$. This fundamentally reshapes our understanding of problem equivalence beyond rank invariance alone. Notably, deceptiveness is absent in one-dimensional landscapes but becomes prevalent in three dimensions---appearing in $70\%$ of classes, often combined with neutrality ($45\%$). Importantly, no one-dimensional or two-dimensional landscape classes combine neutrality with strict deceptiveness (strict suboptima), nor plateaus with deceptiveness (strict or non-strict suboptima). However, three-dimensional landscapes exhibit all facets of difficulty in various combinations, with neutrality, plateaus, and deceptiveness each appearing independently or together. These findings further illustrate the critical role of understanding landscape topology in algorithm performance: we found that a multi-start first-improvement hill climber performs faster than its best-improvement counterpart in almost $60\%$ of three-dimensional classes. The inventory thus provides a deeper understanding of how landscape features interact with algorithm behavior and underscores the importance of algorithm design choices based on landscape features, as the relative performance of hill-climbing strategies is directly tied to the presence of plateaus and local optima.

This complete catalog of minimal landscapes serves as a foundational resource for benchmark design, offering precise building blocks to construct problems with controlled difficulty. By providing all possible ``atoms'' of combinatorial landscapes, the inventory enables a bottom-up approach to problem design and algorithm analysis. Beyond its theoretical value, the inventory allows mapping real-world problems to catalog entries. By enumerating sub-problems of dimension lower than $3$, it enables the empirical study of the probability of occurrence of each class in practical applications. While $12\,007$ unique classes exist, we postulate that their distribution varies significantly across problem domains.

Future work can leverage the inventory to develop algorithms that recognize and exploit invariant structures, particularly when solving subproblems in decomposition or grey-box optimization~\citep{whitley2016}. While higher dimensions ($n>3$) remain theoretically interesting, $n \leq 3$ already captures a broad spectrum of complexity through low-dimensional interactions, making it an ideal basis for constructing larger problems (similar to 3-SAT, but richer) where properties can be controlled through subproblem combinations. We expect the properties of large-scale problems to vary significantly depending on which subproblems are used and how they are combined---a direction we plan to investigate in future work. For example, combining subproblems with specific properties (such as deceptiveness or neutrality) could produce problems with predictable difficulties, supporting algorithm improvement and benchmarking. This study establishes a systematic framework for landscape analysis, bridging theory and practice through complete enumeration of fundamental problem structures.

\bigskip
\paragraph{Supplementary material.}%
Supplementary material, including the inventory and invariant landscape 
class properties, is available at the following URL: \url{https://doi.org/10.5281/zenodo.18492019}.%

\bigskip
\paragraph{Acknowledgments.}%
The authors are grateful to \textbf{Roberto Santana} and \textbf{Francisco Chicano} for their valuable insights into this research.
This work has benefited from the support of the National Research Agency under France 2030, MAIA Project \mbox{ANR-22-EXES-0009}.
\bigskip

\small
\bibliographystyle{apalike}

\providecommand{\MaxMinAntSystem}{{$\cal MAX$--$\cal MIN$} {Ant} {System}}
  \providecommand{\rpackage}[1]{{#1}}
  \providecommand{\softwarepackage}[1]{{#1}}
  \providecommand{\proglang}[1]{{#1}} \providecommand{\BIBdepartment}[1]{{#1},
  }

\newpage
\appendix

\section{Appendix}

\begin{table}[h!]
\caption{Landscape-invariant transformations for 2D hypercube landscapes: translations $t_z$ (left) and rotations $r_\sigma$ (right). The decimal equivalents of binary solutions are given in parentheses.\smallskip}
\label{tab:trans_n2}
\centering%
\begin{tabular}{c|cccc}
\toprule
    &   \multicolumn{4}{c}{\textbf{translation} $t_z$}  \\
    &   \multicolumn{4}{c}{$z$} \\
$x$ & \texttt{00} & \texttt{01} & \texttt{10} & \texttt{11} \\
\midrule
\texttt{00} (0) & \texttt{00} (0) & \texttt{01} (1) & \texttt{10} (2) & \texttt{11} (3) \\
\texttt{01} (1) & \texttt{01} (1) & \texttt{00} (0) & \texttt{11} (3) & \texttt{10} (2) \\
\texttt{10} (2) & \texttt{10} (2) & \texttt{11} (3) & \texttt{00} (0) & \texttt{01} (1) \\
\texttt{11} (3) & \texttt{11} (3) & \texttt{10} (2) & \texttt{01} (1) & \texttt{00} (0) \\
\bottomrule
\end{tabular}
\hspace{3ex}
\begin{tabular}{c|cc}
\toprule
    &   \multicolumn{2}{c}{\textbf{rotation} $r_\sigma$}  \\
    &   \multicolumn{2}{c}{\texttt{$\sigma$}}   \\
$x$ &   $\texttt{01} \mapsto \texttt{01}$   &   $\texttt{01} \mapsto \texttt{10}$   \\
\midrule
    \texttt{00} (0) & \texttt{00} (0) & \texttt{00} (0) \\
    \texttt{01} (1) & \texttt{01} (1) & \texttt{10} (2) \\
    \texttt{10} (2) & \texttt{10} (2) & \texttt{01} (1) \\
    \texttt{11} (3) & \texttt{11} (3) & \texttt{11} (3) \\
\bottomrule
\end{tabular}
\end{table}

\begin{table}[h!]
\caption{Landscape-invariant transformations for 3D hypercube landscapes: translations $t_z$ (top) and rotations $r_\sigma$ (bottom). The decimal equivalents of binary solutions are given in parentheses.\smallskip}
\label{tab:trans_n3}
\centering%
\begin{tabular}{c@{\hspace{1.9ex}}|@{\hspace{1.9ex}}c@{\hspace{1.9ex}}c@{\hspace{1.9ex}}c@{\hspace{1.9ex}}c@{\hspace{1.9ex}}c@{\hspace{1.9ex}}c@{\hspace{1.9ex}}c@{\hspace{1.9ex}}c}
\toprule
    &   \multicolumn{8}{c}{\textbf{translation} $t_z$}  \\
    &   \multicolumn{8}{c}{$z$} \\
$x$ & \texttt{000} & \texttt{001} & \texttt{010} & \texttt{011} & \texttt{100} & \texttt{101} & \texttt{110} & \texttt{111} \\
\midrule
\texttt{000} (0) & \texttt{000} (0) & \texttt{001} (1) & \texttt{010} (2) & \texttt{011} (3) & \texttt{100} (4) & \texttt{101} (5) & \texttt{110} (6) & \texttt{111} (7) \\
\texttt{001} (1) & \texttt{001} (1) & \texttt{000} (0) & \texttt{011} (3) & \texttt{010} (2) & \texttt{101} (5) & \texttt{100} (4) & \texttt{111} (7) & \texttt{110} (6) \\
\texttt{010} (2) & \texttt{010} (2) & \texttt{011} (3) & \texttt{000} (0) & \texttt{001} (1) & \texttt{110} (6) & \texttt{111} (7) & \texttt{100} (4) & \texttt{101} (5) \\
\texttt{011} (3) & \texttt{011} (3) & \texttt{010} (2) & \texttt{001} (1) & \texttt{000} (0) & \texttt{111} (7) & \texttt{110} (6) & \texttt{101} (5) & \texttt{100} (4) \\
\texttt{100} (4) & \texttt{100} (4) & \texttt{101} (5) & \texttt{110} (6) & \texttt{111} (7) & \texttt{000} (0) & \texttt{001} (1) & \texttt{010} (2) & \texttt{011} (3) \\
\texttt{101} (5) & \texttt{101} (5) & \texttt{100} (4) & \texttt{111} (7) & \texttt{110} (6) & \texttt{001} (1) & \texttt{000} (0) & \texttt{011} (3) & \texttt{010} (2) \\
\texttt{110} (6) & \texttt{110} (6) & \texttt{111} (7) & \texttt{100} (4) & \texttt{101} (5) & \texttt{010} (2) & \texttt{011} (3) & \texttt{000} (0) & \texttt{001} (1) \\
\texttt{111} (7) & \texttt{111} (7) & \texttt{110} (6) & \texttt{101} (5) & \texttt{100} (4) & \texttt{011} (3) & \texttt{010} (2) & \texttt{001} (1) & \texttt{000} (0) \\
\bottomrule
\end{tabular}

\bigskip

\begin{tabular}{c@{\hspace{0.86ex}}|@{\hspace{0.86ex}}c@{\hspace{0.86ex}}c@{\hspace{0.86ex}}c@{\hspace{0.86ex}}c@{\hspace{0.86ex}}c@{\hspace{0.86ex}}c}
\toprule
    &   \multicolumn{6}{c}{\textbf{rotation} $r_\sigma$}  \\
    &   \multicolumn{6}{c}{\texttt{$\sigma$}}   \\
$x$ & $\texttt{012} \mapsto \texttt{012}$ & $\texttt{012} \mapsto \texttt{021}$ & $\texttt{012} \mapsto \texttt{102}$ & $\texttt{012} \mapsto \texttt{120}$ & $\texttt{012} \mapsto \texttt{201}$ & $\texttt{012} \mapsto \texttt{210}$ \\
\midrule
    \texttt{000} (0) & \texttt{000} (0) & \texttt{000} (0) & \texttt{000} (0) & \texttt{000} (0) & \texttt{000} (0) & \texttt{000} (0) \\
    \texttt{001} (1) & \texttt{001} (1) & \texttt{001} (1) & \texttt{010} (2) & \texttt{100} (4) & \texttt{010} (2) & \texttt{100} (4) \\
    \texttt{010} (2) & \texttt{010} (2) & \texttt{100} (4) & \texttt{001} (1) & \texttt{001} (1) & \texttt{100} (4) & \texttt{010} (2) \\
    \texttt{011} (3) & \texttt{011} (3) & \texttt{101} (5) & \texttt{011} (3) & \texttt{101} (5) & \texttt{110} (6) & \texttt{110} (6) \\
    \texttt{100} (4) & \texttt{100} (4) & \texttt{010} (2) & \texttt{100} (4) & \texttt{010} (2) & \texttt{001} (1) & \texttt{001} (1) \\
    \texttt{101} (5) & \texttt{101} (5) & \texttt{011} (3) & \texttt{110} (6) & \texttt{110} (6) & \texttt{011} (3) & \texttt{101} (5) \\
    \texttt{110} (6) & \texttt{110} (6) & \texttt{110} (6) & \texttt{101} (5) & \texttt{011} (3) & \texttt{101} (5) & \texttt{011} (3) \\
    \texttt{111} (7) & \texttt{111} (7) & \texttt{111} (7) & \texttt{111} (7) & \texttt{111} (7) & \texttt{111} (7) & \texttt{111} (7) \\
\bottomrule
\end{tabular}
\end{table}

\begin{figure}[h!]
\centering
\includegraphics[width=0.99\textwidth,angle=180]{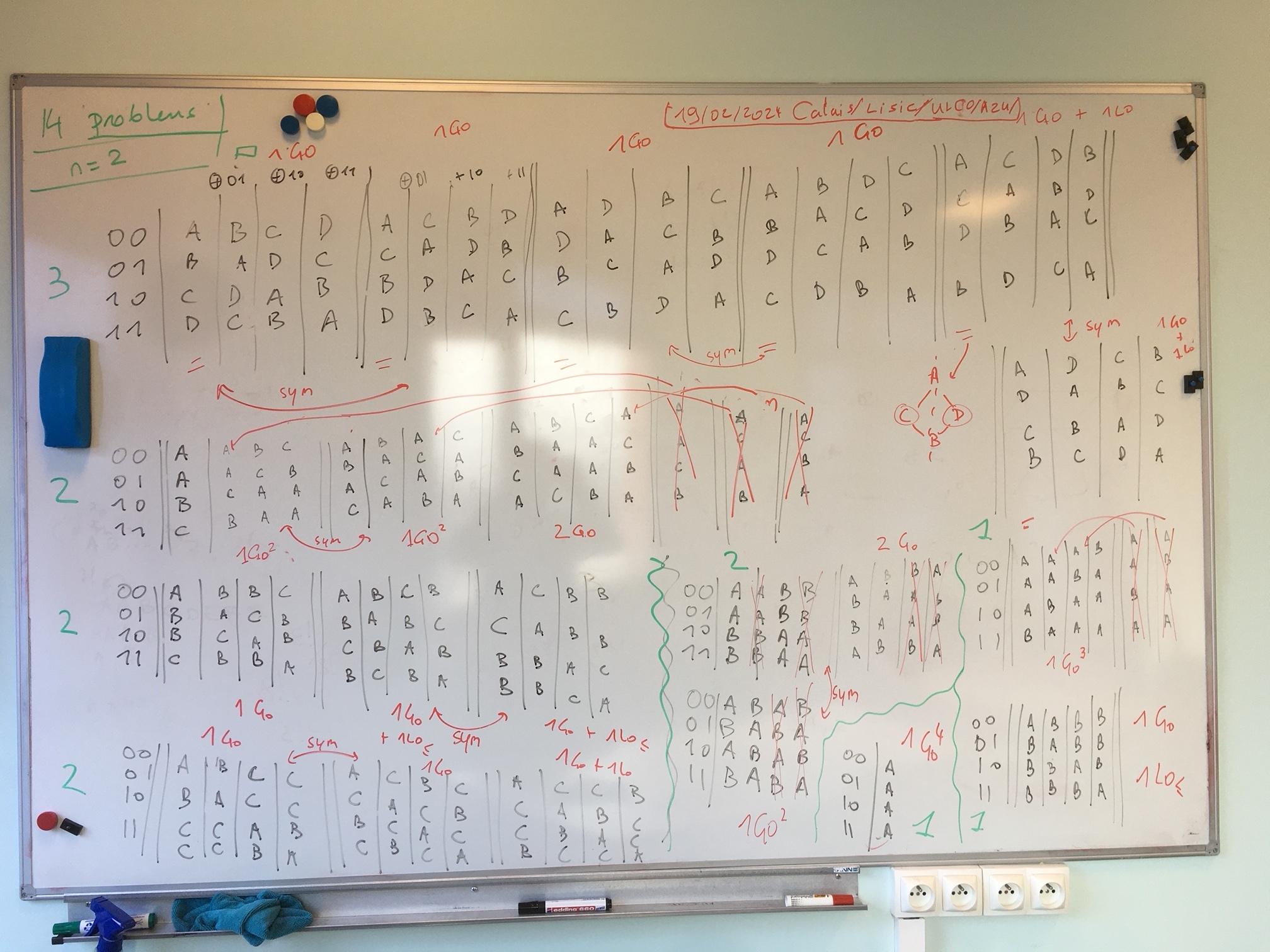}
\caption{Behind the scenes: Original enumeration of 2D invariant landscape classes on the whiteboard.}
\label{fig:a_la_main}
\end{figure}

\begin{table}[h!]
\caption{Detailed properties of 2D landscape classes.\smallskip}
\label{tab:neutrality_n2}
\centering%
\begin{tabular}{r|*{6}{>{\raggedleft\arraybackslash}p{1.6cm}}}
\toprule
\textbf{id} & \textbf{global optima} & \textbf{sub\-optima} & \textbf{neutral networks} & \textbf{optimal plateaus} & \textbf{suboptimal plateaus} & \textbf{neutral degree} \\
\midrule
0 & 1 & 0 & 0 & 0 & 0 & 0 \\
1 & 1 & 0 & 0 & 0 & 0 & 0 \\
2 & 1 & 1 & 0 & 0 & 0 & 0 \\
3 & 2 & 0 & 1 & 1 & 0 & 2 \\
4 & 2 & 0 & 0 & 0 & 0 & 0 \\
5 & 1 & 0 & 0 & 0 & 0 & 0 \\
6 & 1 & 1 & 1 & 0 & 0 & 2 \\
7 & 1 & 0 & 1 & 0 & 0 & 2 \\
8 & 1 & 1 & 0 & 0 & 0 & 0 \\
9 & 3 & 0 & 1 & 1 & 0 & 3 \\
10 & 2 & 0 & 2 & 1 & 0 & 4 \\
11 & 2 & 0 & 0 & 0 & 0 & 0 \\
12 & 1 & 1 & 1 & 0 & 0 & 3 \\
13 & 4 & 0 & 1 & 1 & 0 & 4 \\
\bottomrule
\end{tabular}
\end{table}
 
\begin{table}[h!]
\caption{Detailed expected performance metrics of \textbf{best-improvement} hill climber (single- and multi-start) on 2D landscape classes.\smallskip}
\label{tab:bihc_n2}
\centering%
\begin{tabular}{r|r|rr|rr||r}
\toprule
 & \textbf{success} & \multicolumn{2}{c|}{\textbf{successful trajectories}} & \multicolumn{2}{c||}{\textbf{unsuccessful trajectories}} & \textbf{multi-start}\\
\textbf{id} & \textbf{rate} & \textbf{exp. steps} & \textbf{exp. evals} & \textbf{exp. steps} & \textbf{exp. evals} & \textbf{ERT} \\
\midrule
0  & 1.000 & 1.000 & 5.000 & --    & --    & 5.000 \\
1  & 1.000 & 1.000 & 5.000 & --    & --    & 5.000 \\
2  & 0.750 & 0.667 & 4.333 & 0.000 & 3.000 & 5.333 \\
3  & 1.000 & 0.500 & 4.000 & --    & --    & 4.000 \\
4  & 1.000 & 0.500 & 4.000 & --    & --    & 4.000 \\
5  & 1.000 & 1.000 & 5.000 & --    & --    & 5.000 \\
6  & 0.750 & 0.667 & 4.333 & 0.000 & 3.000 & 5.333 \\
7  & 1.000 & 1.000 & 5.000 & --    & --    & 5.000 \\
8  & 0.750 & 0.667 & 4.333 & 0.000 & 3.000 & 5.333 \\
9  & 1.000 & 0.250 & 3.500 & --    & --    & 3.500 \\
10 & 1.000 & 0.500 & 4.000 & --    & --    & 4.000 \\
11 & 1.000 & 0.500 & 4.000 & --    & --    & 4.000 \\
12 & 0.750 & 0.667 & 4.333 & 0.000 & 3.000 & 5.333 \\
13 & 1.000 & 0.000 & 3.000 & --    & --    & 3.000 \\
\bottomrule
\end{tabular}
\end{table}

\begin{table}[h!]
\caption{Detailed expected performance metrics of \textbf{first-improvement} hill climber (single- and multi-start) on 2D landscape classes, assuming neighbors are explored in random order.\smallskip}
\label{tab:fihc_n2}
\centering%
\begin{tabular}{r|r|rr|rr||r}
\toprule
 & \textbf{success} & \multicolumn{2}{c|}{\textbf{successful trajectories}} & \multicolumn{2}{c||}{\textbf{unsuccessful trajectories}} & \textbf{multi-start}\\
\textbf{id} & \textbf{rate} & \textbf{exp. steps} & \textbf{exp. evals} & \textbf{exp. steps} & \textbf{exp. evals} & \textbf{ERT} \\
\midrule
0  & 1.000 & 1.000 & 4.375 & --    & --    & 4.375 \\
1  & 1.000 & 1.250 & 4.750 & --    & --    & 4.750 \\
2  & 0.500 & 0.500 & 3.500 & 0.500 & 3.500 & 7.000 \\
3  & 1.000 & 0.625 & 3.812 & --    & --    & 3.812 \\
4  & 1.000 & 0.500 & 3.500 & --    & --    & 3.500 \\
5  & 1.000 & 1.000 & 4.375 & --    & --    & 4.375 \\
6  & 0.625 & 0.600 & 3.800 & 0.333 & 3.333 & 5.800 \\
7  & 1.000 & 1.000 & 4.500 & --    & --    & 4.500 \\
8  & 0.500 & 0.500 & 3.500 & 0.500 & 3.500 & 7.000 \\
9  & 1.000 & 0.250 & 3.250 & --    & --    & 3.250 \\
10 & 1.000 & 0.500 & 3.750 & --    & --    & 3.750 \\
11 & 1.000 & 0.500 & 3.500 & --    & --    & 3.500 \\
12 & 0.750 & 0.667 & 4.000 & 0.000 & 3.000 & 5.000 \\
13 & 1.000 & 0.000 & 3.000 & --    & --    & 3.000 \\
\bottomrule
\end{tabular}
\end{table}

\end{document}